\pdfoutput=1
\documentclass[11pt]{article}

\usepackage{EMNLP2023}

\usepackage{times}
\usepackage{latexsym}

\usepackage[T1]{fontenc}
\usepackage[utf8]{inputenc}

\usepackage{microtype}

\usepackage{inconsolata}

\usepackage{tabularx}
\usepackage{bm} % make greek symbol bold
\usepackage{booktabs}
\usepackage{multirow}
\usepackage{amsmath}
\usepackage{graphicx}

\usepackage{tgcursor}
\usepackage[T1]{fontenc}

\usepackage{upquote} % for straight quotation mark '

\usepackage{seqsplit} % split long text

\usepackage[most]{tcolorbox}
\usepackage{etoolbox}
\AtBeginEnvironment{tcolorbox}{\small}

\usepackage{xcolor}
\tcbuselibrary{breakable}

\newcommand{\ralle}{\textsc{RaLLe}}
\newcommand{\rallez}{\textsc{RaLLe }}

\title{\ralle: A Framework for Developing and Evaluating Retrieval-Augmented Large Language Models}

\author{
  Yasuto Hoshi\thanks{\quad These authors contributed equally to this work.}, 
  Daisuke Miyashita\footnotemark[1], 
  Youyang Ng, Kento Tatsuno, \\
  {\bf Yasuhiro Morioka,}
  {\bf Osamu Torii,}
  {\bf Jun Deguchi} \\
  Kioxia Corporation, Japan \\
  \texttt{yasuto1.hoshi@kioxia.com} \\
}

\begin{document}
\maketitle
\begin{abstract}
	Retrieval-augmented large language models (R-LLMs) combine pre-trained large language models (LLMs) with information retrieval systems to improve the accuracy of factual question-answering.
	However, current libraries for building R-LLMs provide high-level abstractions without sufficient transparency for evaluating and optimizing prompts within specific inference processes such as retrieval and generation.
	To address this gap, we present \ralle, an open-source framework designed to facilitate the development, evaluation, and optimization of R-LLMs for knowledge-intensive tasks.
	With \ralle, developers can easily develop and evaluate R-LLMs, improving hand-crafted prompts, assessing individual inference processes, and objectively measuring overall system performance quantitatively.
	By leveraging these features, developers can enhance the performance and accuracy of their R-LLMs in knowledge-intensive generation tasks.
	We open-source our code at \url{https://github.com/yhoshi3/RaLLe}.
\end{abstract}

\section{Introduction}

Large language models (LLMs) have shown great potential for natural language understanding and generation tasks \citep{gpt3,palm,gpt4}.
However, they face challenges when answering factual questions due to hallucinations (or confabulations) \citep{hallucination1,hallucination2}, outdated parametric knowledge \citep{streamingqa}, and memory efficiency of parametric knowledge \citep[e.g.,][]{LMasKB2}.
To address these limitations, researchers have turned to the retrieval-augmented approach used in open-domain question answering (QA) \citep{odqa}, hereinafter referred to as retrieval-augmented LLMs or \textit{R-LLMs}.

In comparison to closed-book settings where language models generate answers without retrieval, R-LLMs (open-book settings) enable the retrieval of relevant information from external databases or corpora \citep{survey,simplyretrieve}, which has led to improved accuracy in open-domain QA \citep{replug}.
Additionally, R-LLMs can acquire extended features even without additional training, such as explicit references, relief from fact hallucination \citep{webgpt}, and easy updates to the knowledge source \citep[e.g.,][]{realm,simplyretrieve}.

Retrieval-augmented generation needs further research and development to reach its full potential.
For example, even though the retriever-reader system has been trained on the Natural Questions (NQ) dataset \citep{nq}, its F1 score on the short answer task is 68.3 and still lags behind the oracle F1 score of 75.7 \citep{unanswerable}.
This implies that further improvements can be made to the retrieval-augmented generation approach.
Additionally, users would be probably aware that the outputs generated by R-LLMs may contain factual errors, particularly when applied to knowledge-intensive tasks.
However, there is currently a lack of accessible evaluation framework to assess their output quality.
This makes it difficult to identify areas for improvement.

\begin{figure*}[t]
	\centering
	\includegraphics[width=1\textwidth]{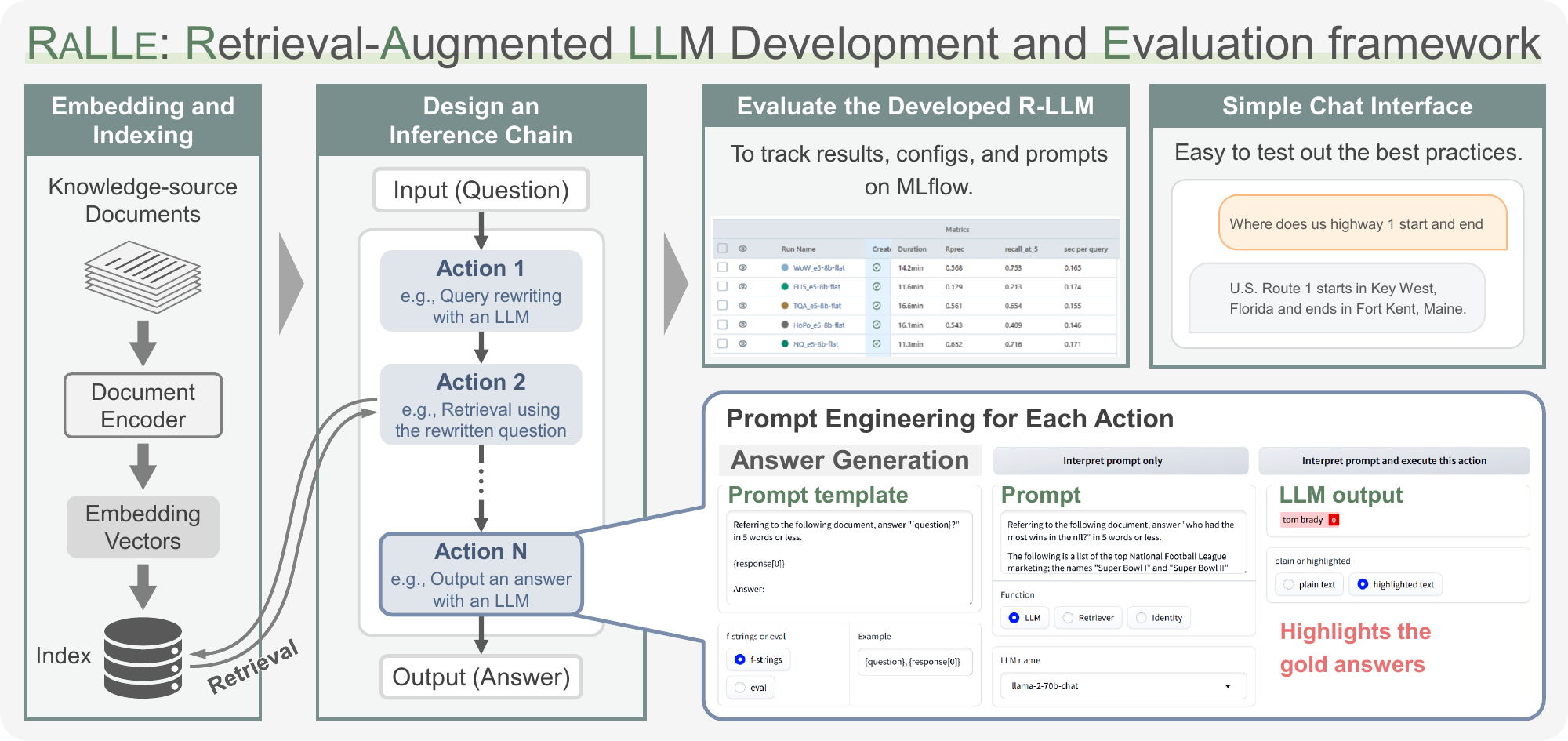}
	\caption{
		Overview of \ralle, our proposed development and evaluation framework for R-LLMs.
		Any number of actions can be defined for an R-LLM.
		Each action can be executed individually to test the corresponding prompts.
		Experimental setup and evaluation results can be tracked using MLflow.
		Additionally, a simple chat interface can be built to test out the best practices from the development and evaluation stages in a practical setting.
	}
	\label{fig:overview}
\end{figure*}

Furthermore, having effective tools for developing R-LLMs is crucial.
These tools should enable the design of inference steps such as retrieve-then-generate, selecting the combination of retrievers and LLMs, evaluating the performance of the entire system, and testing the prompts used in each inference step.
Currently available tools, such as the ChatGPT Retrieval Plugin\footnote{\url{https://github.com/openai/chatgpt-retrieval-plugin}}, Guidance\footnote{\url{https://github.com/microsoft/guidance}}, and LangChain\footnote{Note: Our code does not use either of these.} \citep{langchain}, offer a high degree of abstraction, making it challenging to verify the functionality of individual inference steps or optimize prompts within each step.
This lack of transparency might hinder the optimization of R-LLMs.

In this paper, we propose \ralle, an accessible framework for Retrieval-Augmented Large Language model development and Evaluation.
We also present evaluation results of several R-LLMs that we have constructed by using open-source retrievers and LLMs.
To the best of our knowledge, \rallez is the first framework that empowers R-LLM developers and open-domain QA researchers to efficiently develop, evaluate, and improve R-LLMs using objective metrics.

\rallez offers several key benefits:
\begin{enumerate}
	\item \textit{Easy development and testing}: users can easily select, combine, and test various retrievers and LLMs, especially open-source models, within a graphical interface.
	\item \textit{Objective evaluation of R-LLMs}: \rallez provides reproducible experiments with objective benchmarks/metrics, enabling objective assessments of R-LLM performance.
	\item \textit{Transparent prompt engineering}: all inputs (prompts) and outputs of each action are visible to developers, allowing for easy exploration and optimization of the prompts.
\end{enumerate}

\section{\rallez Usage}\label{sec:usage}

Figure~\ref{fig:overview} presents an overview of the key features of the proposed framework\footnote{Please also review the \href{https://youtu.be/JYbm75qnfTg}{demonstration screencast}.}.
The primary development process involves three stages: (1) embedding and indexing the knowledge source documents, (2) designing an inference chain consisting of an R-LLM with customized prompt templates for each action, and (3) benchmarking the developed R-LLM.

\subsection{Document Embedding and Indexing}\label{sec:embedding}

To begin, the knowledge source documents can be encoded using an arbitrary encoder model, such as a sparse or dense retriever.
For efficient indexing of dense embeddings, several methods are available by default, including Faiss \citep{faiss}, HNSW \citep{hnsw}, and DiskANN \citep{diskann}.
By default, an HNSW index is constructed with $ef\_construction = 128$ (the size of the dynamic list for the nearest neighbors) and $m = 32$ (the number of links created for every new element during graph construction).

\subsection{Chain Construction}\label{sec:chain}

Once the document embedding and indexing are completed, the retrievers (and the corresponding indices) and LLMs can be loaded via the Gradio\footnote{\url{https://www.gradio.app/}}-based GUI \citep{gradio} to establish an inference chain that comprises an R-LLM.
This chain of actions enables users to design a pipeline for multi-step inference, such as $[retrieve]$-$[generate]$, or more intricate workflows such as $[rewrite\ query]$-$[retrieve]$-$[generate]$ proposed in \citet{rewrite}.
The versatility of this feature is especially beneficial in creating the chains tailored to specific use cases.

A single-action chain can function as either a simple retriever that returns the retrieved documents, or a \textit{closed-book} QA that leverages the parametric knowledge of an LLM to provide answers without retrieval.
In contrast, a chain with multiple actions that include retrieval enables retrieval-augmented generation or \textit{open-book} QA, allowing an LLM to access external documents relevant to a question.
Our default setup for R-LLMs consists of two actions: retrieve and generate.

\subsection{Prompt Engineering}\label{sec:prompt_edit}

The \rallez framework allows developers to interactively craft customized prompt templates for LLMs and even for search queries on a per-chain basis.
Each action can be executed independently, enabling precise control over LLM responses, such as specifying the desired output format or suppressing undesirable hallucinations.
To enhance the versatility of prompt development, \rallez integrates support for f-strings and eval() function in Python.

\subsection{Experiment Tracking}\label{sec:mlflow}

We utilize MLflow \citep{mlflow} to track the experiments, along with their associated configuration files and prompt templates.
This allows us to compare the performance of different experiment runs objectively, which enables us to develop even better R-LLMs.

\subsection{Chat AI}

\rallez also provides support for building a simple chat interface.
This enables users to test out best practices from the development and evaluation stages in a practical setting.

\section{Experimental Settings}\label{sec:experiments}

In this section, we evaluate the performance of R-LLMs constructed with several combinations of open-source retrievers and LLMs on knowledge-intensive tasks.

\subsection{Tasks and Datasets}\label{sec:tasks}

We employ KILT (Knowledge Intensive Language Tasks) benchmark \citep{kilt}, an extensive benchmark that encompasses 11 datasets across five knowledge-intensive natural language processing tasks: fact checking, entity linking, slot filling, open-domain question answering, and dialogue (for further details of KILT, see \citet{kilt}).
We use the training sets for developing prompts and the development set for evaluation.

As the knowledge source, we utilize the pre-processed Wikipedia passages provided by KILT.
The passages are derived from English Wikipedia articles based on the 2019/08/01 Wikipedia dump data, consisting of a total of 5.9 million articles and 22.2 million 100-word passages.
For both dense and sparse retrievers, we use the set of 100-word passages after additional pre-processing that prepends the title of the article to each passage.

Note that \rallez is dataset-agnostic, allowing developers to use their own QA datasets and corpora for development and evaluation.
See Appendix~\ref{sec:custom_datasets} for more information.

\subsection{Models}\label{sec:models}

This subsection details the retrievers and LLMs employed to build R-LLMs in our experiments.
\rallez allows practitioners and researchers to easily experiment with the most recent models available in open-source repositories.
With the exception of BM25, all models are available from Hugging Face \citep{huggingface} (see Appendix~\ref{sec:model_info} for the summary).

\subsubsection{LLMs}\label{sec:llms}

The LLM used within the R-LLM must comprehend instructions provided in a prompt and generate appropriate responses based on the given information.
To achieve this, we use instruction-tuned LLMs with a temperature parameter set to zero for optimal performance and reproducibility.

\textbf{Llama-2-chat} is tuned with supervised fine-tuning and reinforcement learning with human feedback (RLHF) \citep{rlhf1,rlhf2} to align to human preferences for helpfulness and safety \citep{llama2}.
In our experiments, we utilize both 13-billion (\textbf{Llama2-13B}) and 70-billion (\textbf{Llama2-70B}) models.

\textbf{WizardVicunaLM-13B\footnote{\url{https://huggingface.co/junelee/wizard-vicuna-13b}} (W-Vicuna-13B)} \citep{wizardvicuna} is formed by combining the concepts of WizardLM \citep{wizardlm} (refining the initial instructions with Evol-Instruct method \citep{wizardlm}) and Vicuna \citep{vicuna} (a fine-tuned LLaMA model \citep{llama} with multi-round conversation data from chatbots).

\subsubsection{Retrievers}\label{sec:retrievers}

We experiment with both sparse and dense retrievers for document retrieval.
Specifically, we select dense retrievers that have achieved high accuracy on the retrieval task of Massive Text Embedding Benchmark (MTEB) \citep{mteb} leaderboard\footnote{\url{https://huggingface.co/spaces/mteb/leaderboard}\label{fot:hf_leaderboard}} as of July 2023.
A list of the retrievers used in our study can be found in Table~\ref{tab:retrievers}.
In the open-book experiments, the top-5 most relevant documents are retrieved.

As the metrics of retrieval performance, we follow \citet{kilt} and use the page-level R-precision \citep{rprec} and recall@5.
The page-level R-precision is the percentage of $R$ gold pages inside each provenance set among the top-$R$ retrieved pages.
Typically, R-Precision is equivalent to Precision@1 except FEVER and HotPotQA (multi-hop datasets).

\begin{table}[t]
	\centering
	\renewcommand{\arraystretch}{1.1}
	{
		\begin{tabular}{lcccc}
			\toprule
			Model & dim.  & max len. & MTEB Retrieval    \\ % & BEIR              \\
			\midrule
			BM25  & -     & -        & 42.3$^\spadesuit$ \\ %& 42.9$^\spadesuit$ \\
			m-e5  & 1,024 & 514      & 51.43             \\ % & -                 \\
			e5    & 1,024 & 512      & 50.56             \\ % & -                 \\
			\bottomrule
		\end{tabular}
	}
	\caption{
		Summary of the retrievers used in our evaluation.
		Dimensions of a dense embedding vector are shown in \textit{dim.}, while the maximum token length of an input sequence is \textit{max len.}.
		The evaluation metric for MTEB Retrieval is nDCG@10.
		$^\spadesuit$: Results from \citet{ram2022you}.
		Results on MTEB Retrieval except BM25 are copied from MTEB leaderboard\footref{fot:hf_leaderboard}.
	}
	\label{tab:retrievers}
\end{table}

\newcommand{\sgr}[1]{\textcolor{lightgray}{#1}}

\begin{table*}[t]
	\centering
	\renewcommand{\arraystretch}{1.2}
	\scalebox{0.65}
	{
		\begin{tabular}{l|ccccccccccc}
			\toprule
			\multicolumn{1}{c}{}                                     & Fact Check.                          & \multicolumn{3}{c}{Entity Linking}            & \multicolumn{2}{c}{Slot Filling}              & \multicolumn{4}{c}{Open Domain QA}            & \multicolumn{1}{c}{Dial.}                                                                                                                                                                                                                                               \\
			\cmidrule(rl){2-2} \cmidrule(rl){3-5} \cmidrule(rl){6-7} \cmidrule(rl){8-11} \cmidrule(rl){12-12}
			\textit{Dataset}                                         & \textbf{FEV}                         & \textbf{AY2}                                  & \textbf{WnWi}                                 & \textbf{WnCw}                                 & \textbf{T-REx}                                & \textbf{zsRE}                        & \textbf{NQ}                                   & \textbf{HoPo}                              & \textbf{TQA}                                  & \textbf{ELI5}    & \textbf{WoW}     \\
			\midrule
			\textit{Model / Metric}                                  & \multicolumn{6}{c|}{Accuracy}        & \multicolumn{3}{c|}{Exact Match}              & \multicolumn{1}{c|}{RL}                       & \multicolumn{1}{c}{F1}                                                                                                                                                                                                                                                                                                  \\
			\midrule
			BART-large$^\diamondsuit$ (\small{\textit{closed-book}}) & \underline{80.7}                     & \textbf{86.6}                                 & 47.9                                          & \textbf{48.0}                                 & \underline{43.8}                              & 3.0                                  & 26.2                                          & 16.9                                       & 32.5                                          & \underline{22.7} & \textbf{13.8}    \\
			Llama2-70B (\small{\textit{closed-book}})                & 33.6 \scriptsize{(74.9)}             & 39.8 \scriptsize{(54.5)}                      & 42.8 \scriptsize{(\textbf{53.8})}             & 39.2 \scriptsize{(\textbf{55.7})}             & 28.5 \scriptsize{(40.5)}                      & 11.3 \scriptsize{(13.6)}             & 19.6 \scriptsize{(37.4)}                      & 13.9 \scriptsize{(25.1)}                   & 67.4 \scriptsize{(80.8)}                      & \textbf{23.0}    & \underline{13.3} \\
			\midrule
			RAG$^\diamondsuit$                                       & \textbf{87.7}                        & \underline{77.4}                              & \textbf{49.0}                                 & \underline{46.7}                              & \textbf{61.5}                                 & \textbf{47.4}                        & \textbf{48.8}                                 & \underline{27.7}                           & 61.7                                          & 16.1             & \underline{13.3} \\
			e5 + W-Vicuna-13B                                        & 10.6 \scriptsize{(42.4)}             & 51.2 \scriptsize{(\textbf{57.9})}             & \underline{48.6} \scriptsize{(51.4)}          & 45.6 \scriptsize{(51.4)}                      & 31.6 \scriptsize{(46.1)}                      & 23.0 \scriptsize{(29.3)}             & 18.7 \scriptsize{(38.0)}                      & 19.7 \scriptsize{(28.3)}                   & 43.1 \scriptsize{(67.7)}                      & 21.4             & 12.3             \\
			e5 + Llama2-13B                                          & 66.3 \scriptsize{(73.5)}             & \sgr{51.2 \scriptsize{(57.9)}}                & \sgr{48.6 \scriptsize{(51.4)}}                & \sgr{45.6 \scriptsize{(51.4)}}                & 17.2 \scriptsize{(42.3)}                      & 31.7 \scriptsize{(41.1)}             & 36.1 \scriptsize{(43.3)}                      & 14.3 \scriptsize{(25.5)}                   & 56.3 \scriptsize{(76.2)}                      & 20.9             & 12.3             \\
			BM25 + Llama2-70B                                        & 46.2 \scriptsize{(86.3)}             & 18.0 \scriptsize{(35.9)}                      & 19.1 \scriptsize{(32.2)}                      & 14.2 \scriptsize{(30.9)}                      & 25.9 \scriptsize{(43.0)}                      & 31.4 \scriptsize{(37.8)}             & 25.3 \scriptsize{(34.3)}                      & 25.9 \scriptsize{(33.4)}                   & 65.8 \scriptsize{(80.0)}                      & 21.3             & 12.2             \\
			\midrule
			e5 + Llama2-70B                                          & 49.9 \scriptsize{(\textbf{88.6})}    & \sgr{51.2 \scriptsize{(57.9)}}                & \sgr{48.6 \scriptsize{(51.4)}}                & \sgr{45.6 \scriptsize{(51.4)}}                & 28.9 \scriptsize{(49.2)}                      & 35.0  \scriptsize{(\textbf{43.2})}   & \underline{36.4} \scriptsize{(\textbf{48.8})} & \textbf{28.1} \scriptsize{(\textbf{35.8})} & \underline{71.1} \scriptsize{(\textbf{83.9})} & 21.5             & 13.2             \\
			\quad \texttt{e5 (DiskANN)}                              & 49.9 \scriptsize{(87.9)}             & 44.3 \scriptsize{(50.5)}                      & 45.3 \scriptsize{(48.1)}                      & 43.0 \scriptsize{(48.8)}                      & 25.3  \scriptsize{(43.9)}                     & 32.1 \scriptsize{(37.9)}             & 36.1  \scriptsize{(48.4)}                     & 26.7 \scriptsize{(34.3)}                   & 70.4 \scriptsize{(83.2)}                      & 21.5             & 13.1             \\
			\quad \texttt{top-2}                                     & 49.3 \scriptsize{(88.1)}             & \sgr{51.2 \scriptsize{(57.9)}}                & \sgr{48.6 \scriptsize{(51.4)}}                & \sgr{45.6 \scriptsize{(51.4)}}                & 23.5 \scriptsize{(44.9)}                      & 34.7 \scriptsize{(43.0)}             & 33.7 \scriptsize{(46.2)}                      & 23.8 \scriptsize{(34.2)}                   & \textbf{71.3} \scriptsize{(82.9)}             & 21.6             & \underline{13.3} \\
			\quad \texttt{top-10}                                    & 50.2 \scriptsize{(88.0)}             & \sgr{51.2 \scriptsize{(57.9)}}                & \sgr{48.6 \scriptsize{(51.4)}}                & \sgr{45.6 \scriptsize{(51.4)}}                & 31.1 \scriptsize{(\textbf{49.3})}             & \underline{35.4} \scriptsize{(42.5)} & 35.2 \scriptsize{(48.1)}                      & 24.9 \scriptsize{(35.7)}                   & 59.3 \scriptsize{(82.8)}                      & 21.5             & 13.2             \\
			\midrule
			\textit{Model / Metric}                                  & \multicolumn{6}{c|}{KILT-Accuracy}   & \multicolumn{3}{c|}{KILT-EM}                  & \multicolumn{1}{c|}{KILT-RL}                  & \multicolumn{1}{c}{KILT-F1}                                                                                                                                                                                                                                                                                             \\
			\midrule
			RAG$^\diamondsuit$                                       & \textbf{55.5}                        & \textbf{77.4}                                 & \textbf{49.0}                                 & \textbf{46.7}                                 & \textbf{25.4}                                 & \textbf{42.6}                        & \textbf{36.3}                                 & 3.1                                        & 36.1                                          & \textbf{2.7}     & 7.5              \\
			e5 + W-Vicuna-13B                                        & 8.4 \scriptsize{(33.5)}              & \underline{51.2} \scriptsize{(\textbf{51.2})} & \underline{48.6} \scriptsize{(\textbf{48.6})} & \underline{45.5} \scriptsize{(\textbf{45.5})} & 19.0 \scriptsize{(28.0)}                      & 22.2 \scriptsize{(28.1)}             & 14.4 \scriptsize{(27.8)}                      & 8.6 \scriptsize{(11.9)}                    & 26.6 \scriptsize{(40.3)}                      & \textbf{2.7}     & 7.3              \\
			e5 + Llama2-13B                                          & \underline{53.1} \scriptsize{(58.7)} & \sgr{51.2 \scriptsize{(51.2)}}                & \sgr{48.6 \scriptsize{(48.6)}}                & \sgr{45.5 \scriptsize{(45.5)}}                & 11.5 \scriptsize{(25.7)}                      & 29.8 \scriptsize{(38.5)}             & 27.5 \scriptsize{(32.5)}                      & 5.6 \scriptsize{(10.6)}                    & 34.7 \scriptsize{(46.1)}                      & \textbf{2.7}     & 7.4              \\
			BM25 + Llama2-70B                                        & 21.9 \scriptsize{(44.4)}             & 17.6 \scriptsize{(17.6)}                      & 18.9 \scriptsize{(18.9)}                      & 13.9 \scriptsize{(13.9)}                      & 14.5 \scriptsize{(22.5)}                      & 24.9 \scriptsize{(29.6)}             & 9.3 \scriptsize{(12.4)}                       & 4.5 \scriptsize{(5.9)}                     & 23.6 \scriptsize{(27.9)}                      & \underline{1.5}  & 4.0              \\
			\midrule
			e5 + Llama2-70B                                          & 40.2  \scriptsize{(\textbf{71.2})}   & \sgr{51.2 \scriptsize{(51.2)}}                & \sgr{48.6 \scriptsize{(48.6)}}                & \sgr{45.5 \scriptsize{(45.5)}}                & 19.2 \scriptsize{(29.7)}                      & 32.8 \scriptsize{(40.4)}             & \underline{27.7} \scriptsize{(\textbf{36.3})} & \textbf{11.3} \scriptsize{(\textbf{14.5})} & \underline{42.8} \scriptsize{(\textbf{49.7})} & \textbf{2.7}     & \underline{8.1}  \\
			\quad e5 (DiskANN)                                       & 38.3 \scriptsize{(68.5)}             & 44.3 \scriptsize{(44.3)}                      & 45.3 \scriptsize{(45.3)}                      & 42.8 \scriptsize{(42.8)}                      & 19.3 \scriptsize{(24.2)}                      & 30.2 \scriptsize{(35.5)}             & 27.3 \scriptsize{(35.9)}                      & 9.3 \scriptsize{(12.1)}                    & 42.1 \scriptsize{(49.0)}                      & \textbf{2.7}     & 8.0              \\
			\quad \texttt{top-2}                                     & 39.6 \scriptsize{(70.7)}             & \sgr{51.2 \scriptsize{(51.2)}}                & \sgr{48.6 \scriptsize{(48.6)}}                & \sgr{45.5 \scriptsize{(45.5)}}                & 15.6 \scriptsize{(28.0)}                      & 32.9 \scriptsize{(\textbf{40.6})}    & 25.7 \scriptsize{(35.2)}                      & 7.6 \scriptsize{(13.1)}                    & \textbf{43.1} \scriptsize{(49.3)}             & \textbf{2.7}     & \textbf{8.3}     \\
			\quad \texttt{top-10}                                    & 40.4 \scriptsize{(70.7)}             & \sgr{51.2 \scriptsize{(51.2)}}                & \sgr{48.6 \scriptsize{(48.6)}}                & \sgr{45.5 \scriptsize{(45.5)}}                & \underline{20.5} \scriptsize{(\textbf{29.9})} & \underline{33.2} \scriptsize{(39.8)} & 27.1 \scriptsize{(36.1)}                      & \underline{9.9} \scriptsize{(14.3)}        & 36.1 \scriptsize{(48.9)}                      & \textbf{2.7}     & \underline{8.1}  \\
			\bottomrule
		\end{tabular}
	}
	\caption{
		Downstream performance on KILT dev set.
		Following \citet{kilt}, we report the results of typical metrics for each dataset, with bold indicating the best result and underlined indicating the second.
		The metrics with the prefix \textit{KILT-} award output performance only when R-Prec = 1 (retrieval success).
		The figures in parentheses represent has\_answer percentage, which corresponds to the proportion of questions with gold answers included in the final output.
		The figures shown in gray are copied from the column above because they do not change based on the given setting (we use the Identity function of \rallez for the tasks, rather than an LLM).
		% See text for details.
		$^\diamondsuit$: Results from \citet{kilt}.
	}
	\label{tab:kilt_downstream_open}
\end{table*}

\textbf{BM25} \citep{bm25} is a bag-of-words retrieval function based on the term-matching.
We use the Pyserini \citep{pyserini} implementation of unigram BM25 with the default parameters of $k_1 = 0.9$ (term frequency scaling) and $b = 0.4$ (document length normalization).
The documents for BM25 retrieval is the same 100-word passages as the dense retrievers.

\textbf{e5-large-v2}\footnote{\url{https://huggingface.co/intfloat/e5-large-v2}} (\textbf{e5}) \citep{e5} is a supervised bi-encoder model with a query encoder and a document encoder.
\textbf{multilingual-e5-large}\footnote{\url{https://huggingface.co/intfloat/multilingual-e5-large}} (\textbf{m-e5}) is a multilingual fine-tuned e5 model.

\subsection{Prompts}\label{sec:prompts}

We utilize custom-designed prompt templates that are specifically crafted for each dataset in KILT.
\rallez accepts templates with non-natural language formats, such as f-strings and eval() functions in Python.
This allows developers to carefully craft their prompt templates for optimal performance.
The prompt templates used in our experiments are shown in Appendix~\ref{sec:our_prompts}.

For entity linking task of KILT (AY2, WnWi, and WnCw), we employ a \texttt{REWRITE-EL} template by default for search queries.
This template extracts the specific entity mentions being questioned as a query, as employing an entire span of a question is unlikely to find relevant documents (we will discuss in Section~\ref{sec:kilt_retrieval}).
After retrieving the relevant documents, the top-1 Wikipedia title is output as an answer.
As a result, the downstream accuracy in entity linking task is not affected by the number of retrieved documents (if one or more).

\section{KILT Benchmark Results}\label{sec:kilt_results}

This section provides the downstream and retrieval performance of the R-LLMs developed and evaluated using \ralle.

\subsection{Baseline}\label{sec:kilt_baseline}

We compare our results with those of the BART-large model \citep{bart} for the closed-book setting and the RAG model \citep{rag} for the open-book setting, which presented in \citet{kilt}.
Notably, these baseline models were specifically fine-tuned on the KILT benchmark, whereas our chosen LLMs and constructed R-LLMs were not.
See also Appendix \ref{sec:rag_details} for additional information of the baselines.

\begin{table*}[t]
	\centering
	\renewcommand{\arraystretch}{1.1}
	\scalebox{0.708}
	{
		\begin{tabular}{l|ccccccccccc|l}
			\toprule
			\multicolumn{1}{c}{}          & Fact Check.                              & \multicolumn{3}{c}{Entity Linking} & \multicolumn{2}{c}{Slot Filling} & \multicolumn{4}{c}{Open Domain QA} & \multicolumn{1}{c}{Dial.} & \multicolumn{1}{c}{}                                                                                                                \\
			\cmidrule(rl){2-2} \cmidrule(rl){3-5} \cmidrule(rl){6-7} \cmidrule(rl){8-11} \cmidrule(rl){12-12}
			\textit{Dataset}              & \textbf{FEV}                             & \textbf{AY2}                       & \textbf{WnWi}                    & \textbf{WnCw}                      & \textbf{T-REx}            & \textbf{zsRE}        & \textbf{NQ}   & \textbf{HoPo} & \textbf{TQA}  & \textbf{ELI5} & \textbf{WoW}  & Avg.                         \\
			\midrule
			\textit{Model}                & \multicolumn{11}{c}{\large{R-Precision}} & \multicolumn{1}{c}{}                                                                                                                                                                                                                                                         \\
			\midrule
			RAG$^\diamondsuit$            & 63.5                                     & \textbf{77.4}                      & 49.0                             & 46.7                               & 29.3                      & 65.4                 & 60.3          & 30.8          & 49.3          & \textbf{16.4} & 46.7          & 48.6                         \\
			BM25                          & 52.1                                     & 17.7                               & 20.6                             & 15.3                               & 34.0                      & 57.7                 & 26.3          & 41.3          & 31.7          & 6.8           & 28.8          & 30.2                         \\
			\quad $-$ \texttt{REWRITE-EL} & \sgr{52.1}                               & 3.0 \scriptsize{($-$14.7)}         & 0.1 \scriptsize{($-$20.5)}       & 2.8* \scriptsize{($-$12.5)}        & \sgr{34.0}                & \sgr{57.7}           & \sgr{26.3}    & \sgr{41.3}    & \sgr{31.7}    & \sgr{6.8}     & \sgr{28.8}    & 25.9 \scriptsize{($-$4.3)}   \\
			m-e5 (Flat)                   & 81.7                                     & 41.8                               & 45.8                             & 41.6                               & \textbf{47.1}             & 81.4                 & 63.0          & 54.0          & \textbf{56.1} & 11.9          & \textbf{57.9} & 52.9                         \\
			\quad $-$ \texttt{REWRITE-EL} & \sgr{81.7}                               & 3.2 \scriptsize{($-$38.6)}         & 0.1 \scriptsize{($-$45.7)}       & 3.1 \scriptsize{($-$38.5)}         & \sgr{47.1}                & \sgr{81.4}           & \sgr{63.0}    & \sgr{54.0}    & \sgr{56.1}    & \sgr{11.9}    & \sgr{57.9}    & 41.8 \scriptsize{($-$11.1)}  \\
			% m-e5 (HNSW)                 & 57.0                            & 2.0                                & 0.1                                & 1.3                                & 23.3                      & 45.5                 & 50.7          & 28.4          & 42.5          & 10.0          & 52.8          & 28.5                               \\
			e5 (Flat)                     & \textbf{82.0}                            & 51.6                               & \textbf{51.6}                    & \textbf{49.2}                      & 45.3                      & \textbf{81.9}        & \textbf{65.2} & \textbf{54.3} & \textbf{56.1} & 12.9          & 56.8          & \textbf{55.2}                \\
			\quad $-$ \texttt{REWRITE-EL} & \sgr{82.0}                               & 3.4 \scriptsize{($-$48.2)}         & 0.0 \scriptsize{($-$51.6)}       & 2.6 \scriptsize{($-$46.6)}         & \sgr{45.3}                & \sgr{81.9}           & \sgr{65.2}    & \sgr{54.3}    & \sgr{56.1}    & \sgr{12.9}    & \sgr{56.8}    & 41.9 \scriptsize{($-$13.3)}  \\
			e5 (HNSW)                     & 67.9                                     & 38.9                               & 42.3                             & 40.5                               & 23.1                      & 53.0                 & 60.3          & 34.9          & 50.4          & 10.2          & 54.5          & 43.3                         \\
			\quad $-$ \texttt{REWRITE-EL} & \sgr{67.9}                               & 2.9 \scriptsize{($-$36.0)}         & 0.0 \scriptsize{($-$42.3)}       & 1.6 \scriptsize{($-$38.9)}         & \sgr{23.1}                & \sgr{53.0}           & \sgr{60.3}    & \sgr{34.9}    & \sgr{50.4}    & \sgr{10.2}    & \sgr{54.5}    & 32.6  \scriptsize{($-$10.7)} \\
			e5 (DiskANN)                  & 78.8                                     & 44.7                               & 47.8                             & 46.0                               & 37.1                      & 74.5                 & 64.9          & 49.1          & 55.4          & 12.9          & 56.6          & 51.6                         \\
			\quad $-$ \texttt{REWRITE-EL} & \sgr{78.8}                               & 3.2 \scriptsize{($-$41.5)}         & 0.1 \scriptsize{($-$47.7)}       & 1.8 \scriptsize{($-$44.2)}         & \sgr{37.1}                & \sgr{74.5}           & \sgr{64.9}    & \sgr{49.1}    & \sgr{55.4}    & \sgr{12.9}    & \sgr{56.6}    & 39.4 \scriptsize{($-$12.2)}  \\
			\bottomrule
		\end{tabular}
	}
	\caption{
		Retrieval performances on KILT dev set.
		We report page-level R-Precision on KILT development set.
		\textit{Avg.} refers to macro-average of the retrieval scores in each dataset.
		Bold indicates the best result.
		$^\diamondsuit$: Results from \citet{kilt}.
		*: BM25 (without \texttt{REWRITE-EL}) failed with long queries (45 out of 5,599 questions) in WnCw.
		% Note that R-precision would not change for top-2 or top-5 retrieval by definition, except in cases of multi-hop questions (FEV and HoPo).
	}
	\label{tab:kilt_retrieval}
\end{table*}

\subsection{Downstream Performance}\label{sec:kilt_downstream}

We summarize the downstream performance\footnote{See also Table \ref{tab:kilt_downstream_closed} in Appendix~\ref{sec:kilt_downstream_closed} for additional results in a closed-book setting.} in Table \ref{tab:kilt_downstream_open}.
\rallez also includes has\_answer percentage for short answers, a proxy metric to measure the proportion of questions that contain gold answers within the final output generated by an R-LLM (see Appendix~\ref{sec:has_answer} for more details).

Our constructed R-LLM (e5 + Llama2-70B) surpasses the performance of the RAG model on both HoPo and TQA, despite not being fine-tuned with KILT like RAG.
Moreover, our constructed R-LLMs demonstrate acceptable accuracy levels on other datasets as well, without any significant drawbacks.
The results indicate that the LLMs used in this study exhibit certain ability to comprehend the retrieved documents.

Furthermore, our analysis reveals several factors that could contribute to improvement of downstream performance, including retrieval augmentation (except ELI5), increased model scale (except FEV and T-REx), and referring to more documents during generation (except NQ, HoPo, TQA and WoW).
However, some datasets exhibits exceptions to these tendencies or had lower performance compared to their corresponding has\_answer percentage (such as FEV, T-REx, NQ, and TQA).
To address this issue, developers can improve the R-LLM with \rallez by refining the inference chain and the prompt templates.
In Section~\ref{sec:3_action}, we provide our initial attempts at developing inference chains with three actions on several datasets.

Overall, the downstream evaluation results provide valuable insights into how well the constructed R-LLMs perform on knowledge-intensive tasks, enabling developers to identify areas for improvement.

\subsection{Retrieval Performance}\label{sec:kilt_retrieval}

Table~\ref{tab:kilt_retrieval} shows retrieval performance of the chosen retrievers on KILT development set (see also Table~\ref{tab:recall} in Appendix for the results of recall@5).
According to Table~\ref{tab:kilt_retrieval}, e5 (with Faiss Flat index) achieves the highest retrieval performance on average, though m-e5 is better on MTEB Retrieval task (Table \ref{tab:retrievers}).
Despite the superior retrieval accuracy of e5 compared to RAG on KILT, the downstream performance of the R-LLM which employs e5 falls short of that of RAG (Table \ref{tab:kilt_downstream_open}).
This indicates that there is potential room for improvement through further optimized prompts to enhance the performance on a target dataset.

As described in Section~\ref{sec:prompts}, \texttt{REWRITE-EL} serves as the default template for search queries related to entity linking task (AY2, WnWi, and WnCw).
As shown in Table~\ref{tab:kilt_retrieval}, employing the \texttt{REWRITE-EL} template leads to higher retrieval accuracy when compared to using the full question text as a search query ($-$ \texttt{REWRITE-EL} setting).
This indicates that omitting unnecessary information from the search queries is helpful especially for entity linking task.

\begin{table}[t]
	\renewcommand{\arraystretch}{1.15}
	\scalebox{0.8}
	{
		\begin{tabular}{p{3.5cm}ccc}
			\toprule
			\multicolumn{4}{c}{\textit{Retrieval}}                                                             \\
			\midrule
			Model                                                              & Avg. R-Prec & Memory  & sec/Q \\
			\hline
			BM25                                                               & 30.2        & -       & 0.121 \\
			e5 (Flat)                                                          & 55.2        & 84.8 GB & 0.169 \\
			e5 (HNSW)                                                          & 43.3        & 90.4 GB & 0.008 \\
			e5 (DiskANN)                                                       & 51.6        & 10.9 GB & 0.022 \\
			\midrule
			\multicolumn{3}{c}{\textit{Completion in the Closed-Book Setting}} & sec/Q                         \\
			\midrule
			Llama-70B                                                          &             &         & 6.727 \\
			\midrule
			\multicolumn{3}{c}{\textit{Retrieval + Generation}}                & sec/Q                         \\
			\midrule
			\multicolumn{3}{l}{BM25 + Llama2-70B}                              & 3.637                         \\
			\multicolumn{3}{l}{e5 + Llama2-70B}                                & 3.793                         \\
			\multicolumn{3}{l}{e5 (DiskANN) + Llama2-70B}                      & 3.628                         \\
			\bottomrule
		\end{tabular}
	}
	\caption{
		Execution latency in seconds per question (sec/Q).
		\textit{Memory} in Retrieval indicates the maximum (DRAM) memory footprints.
	}
	\label{tab:speed}
\end{table}

\subsection{Speed Analysis}\label{sec:speed}

\rallez allows users to optimize the trade-off between latency (in seconds per question) and accuracy by comparing various configurations.
As demonstrated in Table~\ref{tab:speed}, employing approximate nearest neighbor search (ANNS) algorithms such as HNSW and DiskANN can significantly reduce retrieval latency at the cost of decreased accuracy.
Note that, the optimal balance between speed and accuracy depends on the specific requirements of the application, and \rallez enables users to easily experiment with diverse ANNS settings to determine their impact on both factors.

Notably, DiskANN achieves an accuracy that is only slightly lower than Faiss flat index while significantly improving search speeds, despite requiring less memory footprints than both flat and HNSW indices.
Though the reduction in R-LLM execution time achieved through ANNS may appear relatively minor, the significantly lower DRAM requirements of DiskANN could make it a more practical solution for scenarios where DRAM capacity is limited and the flat index exceeds available DRAM capacity.
For further details regarding latency, refer to Table~\ref{tab:details_speed} in Appendix~\ref{sec:details_speed}.

\section{Conclusion}

This paper introduces \ralle, an accessible framework for developing and evaluating R-LLMs.
We also report evaluation results of several R-LLMs built using open-source retrievers and LLMs on knowledge-intensive tasks.
Overall, \rallez offers a significant advancement in retrieval-augmented generation research, enabling efficient development, evaluation, and improvement of R-LLMs.
We hope that \rallez will contribute to the development of best practices for R-LLMs.

\section*{Limitations}\label{sec:limitations}

All KILT evaluations presented in this paper were conducted using a development set to maintain fairness and consistency across evaluations, as the answers of the test set remain confidential\footnote{\url{https://eval.ai/web/challenges/challenge-page/689/overview}}.

While R-LLMs exhibit high validity, it falls behind the smaller yet specialized model, RAG, on the KILT downstream task (refer to Table \ref{tab:kilt_downstream_open}).
This disparity can be attributed to various factors, including prompt maturity and the ability of LLMs to generate responses.
Although the employed prompts were carefully developed, it is likely that more optimal prompts exist (discussed in Section~\ref{sec:kilt_retrieval}).
Moreover, fine-tuning LLMs with retrieval-augmented generation tasks might enhance their performance on downstream tasks.
Therefore, the evaluation accuracy reported herein would represent a conservative estimate.

Prompt engineering is a crucial aspect of the retrieval-augmented generation process, as the generated outputs can differ significantly between models, even when provided with the same prompt.
\rallez offers an advantage in this regard, allowing users to effortlessly experiment with diverse prompts for varying behaviors, datasets, and intricate chain of actions.

In the realm of prompt development, techniques like Automatic Prompt Engineer (APE) \citep{ape} automate the creation of prompts from input-output pairs and sampling to identify the most effective prompts.
However, the input-output pairs in retrieval-augmented generation are distinctly different from those of the simple instruction induction tasks.
Because the input text for retrieval-augmented generation can often be lengthy and complex, it is difficult to automatically induce the effective prompts from the input-output pairs.

This tool enables developers to construct an inference chain with predefined actions, while recent advances have also introduced methods allowing LLMs to determine the actions \citep{react}.
One approach entails retrieving documents using a query rewritten by an LLM and then summarizing them until the desired information is obtained.
However, in our initial experiments (not described in this paper), we observed instances where relatively small LLMs (typically less than 100 billion parameters) became trapped in cycles of repeated retrieval and summarization, hindering their ability to reach the final answer generation.
Our tool addresses this issue by intentionally building explicit inference chains to avoid unintended operations.

\bibliography{custom}

\begin{thebibliography}{42}
\expandafter\ifx\csname natexlab\endcsname\relax\def\natexlab#1{#1}\fi

\bibitem[{Abid et~al.(2019)Abid, Abdalla, Abid, Khan, Alfozan, and Zou}]{gradio}
Abubakar Abid, Ali Abdalla, Ali Abid, Dawood Khan, Abdulrahman Alfozan, and James Zou. 2019.
\newblock \href {http://arxiv.org/abs/1906.02569} {Gradio: Hassle-free sharing and testing of ml models in the wild}.
\newblock \emph{arXiv preprint arXiv:1906.02569}.

\bibitem[{Asai and Choi(2021)}]{unanswerable}
Akari Asai and Eunsol Choi. 2021.
\newblock \href {https://doi.org/10.18653/v1/2021.acl-long.118} {Challenges in information-seeking {QA}: Unanswerable questions and paragraph retrieval}.
\newblock In \emph{Proceedings of the 59th Annual Meeting of the Association for Computational Linguistics and the 11th International Joint Conference on Natural Language Processing (Volume 1: Long Papers)}, pages 1492--1504, Online. Association for Computational Linguistics.

\bibitem[{Bang et~al.(2023)Bang, Cahyawijaya, Lee, Dai, Su, Wilie, Lovenia, Ji, Yu, Chung et~al.}]{hallucination1}
Yejin Bang, Samuel Cahyawijaya, Nayeon Lee, Wenliang Dai, Dan Su, Bryan Wilie, Holy Lovenia, Ziwei Ji, Tiezheng Yu, Willy Chung, et~al. 2023.
\newblock \href {http://arxiv.org/abs/2302.04023} {A multitask, multilingual, multimodal evaluation of chatgpt on reasoning, hallucination, and interactivity}.
\newblock \emph{arXiv preprint arXiv:2302.04023}.

\bibitem[{Borji(2023)}]{hallucination2}
Ali Borji. 2023.
\newblock \href {http://arxiv.org/abs/2302.03494} {A categorical archive of {ChatGPT} failures}.
\newblock \emph{arXiv preprint arXiv:2302.03494}.

\bibitem[{Brown et~al.(2020)Brown, Mann, Ryder, Subbiah, Kaplan, Dhariwal, Neelakantan, Shyam, Sastry, Askell, Agarwal, Herbert-Voss, Krueger, Henighan, Child, Ramesh, Ziegler, Wu, Winter, Hesse, Chen, Sigler, Litwin, Gray, Chess, Clark, Berner, McCandlish, Radford, Sutskever, and Amodei}]{gpt3}
Tom Brown, Benjamin Mann, Nick Ryder, Melanie Subbiah, Jared~D Kaplan, Prafulla Dhariwal, Arvind Neelakantan, Pranav Shyam, Girish Sastry, Amanda Askell, Sandhini Agarwal, Ariel Herbert-Voss, Gretchen Krueger, Tom Henighan, Rewon Child, Aditya Ramesh, Daniel Ziegler, Jeffrey Wu, Clemens Winter, Chris Hesse, Mark Chen, Eric Sigler, Mateusz Litwin, Scott Gray, Benjamin Chess, Jack Clark, Christopher Berner, Sam McCandlish, Alec Radford, Ilya Sutskever, and Dario Amodei. 2020.
\newblock \href {https://proceedings.neurips.cc/paper_files/paper/2020/file/1457c0d6bfcb4967418bfb8ac142f64a-Paper.pdf} {Language models are few-shot learners}.
\newblock In \emph{Advances in Neural Information Processing Systems}, volume~33, pages 1877--1901. Curran Associates, Inc.

\bibitem[{Chase(2023)}]{langchain}
Harrison Chase. 2023.
\newblock {LangChain}.
\newblock \url{https://langchain.com/}.

\bibitem[{Chen et~al.(2017)Chen, Fisch, Weston, and Bordes}]{odqa}
Danqi Chen, Adam Fisch, Jason Weston, and Antoine Bordes. 2017.
\newblock \href {https://doi.org/10.18653/v1/P17-1171} {Reading {W}ikipedia to answer open-domain questions}.
\newblock In \emph{Proceedings of the 55th Annual Meeting of the Association for Computational Linguistics (Volume 1: Long Papers)}, pages 1870--1879, Vancouver, Canada. Association for Computational Linguistics.

\bibitem[{Chiang et~al.(2023)Chiang, Li, Lin, Sheng, Wu, Zhang, Zheng, Zhuang, Zhuang, Gonzalez, Stoica, and Xing}]{vicuna}
Wei-Lin Chiang, Zhuohan Li, Zi~Lin, Ying Sheng, Zhanghao Wu, Hao Zhang, Lianmin Zheng, Siyuan Zhuang, Yonghao Zhuang, Joseph~E. Gonzalez, Ion Stoica, and Eric~P. Xing. 2023.
\newblock \href {https://lmsys.org/blog/2023-03-30-vicuna/} {Vicuna: An open-source chatbot impressing {GPT-4} with 90\%* {ChatGPT} quality}.

\bibitem[{Chowdhery et~al.(2022)Chowdhery, Narang, Devlin, Bosma, Mishra, Roberts, Barham, Chung, Sutton, Gehrmann, Schuh, Shi, Tsvyashchenko, Maynez, Rao, Barnes, Tay, Shazeer, Prabhakaran, Reif, Du, Hutchinson, Pope, Bradbury, Austin, Isard, Gur-Ari, Yin, Duke, Levskaya, Ghemawat, Dev, Michalewski, Garcia, Misra, Robinson, Fedus, Zhou, Ippolito, Luan, Lim, Zoph, Spiridonov, Sepassi, Dohan, Agrawal, Omernick, Dai, Pillai, Pellat, Lewkowycz, Moreira, Child, Polozov, Lee, Zhou, Wang, Saeta, Diaz, Firat, Catasta, Wei, Meier-Hellstern, Eck, Dean, Petrov, and Fiedel}]{palm}
Aakanksha Chowdhery, Sharan Narang, Jacob Devlin, Maarten Bosma, Gaurav Mishra, Adam Roberts, Paul Barham, Hyung~Won Chung, Charles Sutton, Sebastian Gehrmann, Parker Schuh, Kensen Shi, Sasha Tsvyashchenko, Joshua Maynez, Abhishek Rao, Parker Barnes, Yi~Tay, Noam Shazeer, Vinodkumar Prabhakaran, Emily Reif, Nan Du, Ben Hutchinson, Reiner Pope, James Bradbury, Jacob Austin, Michael Isard, Guy Gur-Ari, Pengcheng Yin, Toju Duke, Anselm Levskaya, Sanjay Ghemawat, Sunipa Dev, Henryk Michalewski, Xavier Garcia, Vedant Misra, Kevin Robinson, Liam Fedus, Denny Zhou, Daphne Ippolito, David Luan, Hyeontaek Lim, Barret Zoph, Alexander Spiridonov, Ryan Sepassi, David Dohan, Shivani Agrawal, Mark Omernick, Andrew~M. Dai, Thanumalayan~Sankaranarayana Pillai, Marie Pellat, Aitor Lewkowycz, Erica Moreira, Rewon Child, Oleksandr Polozov, Katherine Lee, Zongwei Zhou, Xuezhi Wang, Brennan Saeta, Mark Diaz, Orhan Firat, Michele Catasta, Jason Wei, Kathy Meier-Hellstern, Douglas Eck, Jeff Dean, Slav Petrov, and Noah Fiedel. 2022.
\newblock \href {https://arxiv.org/abs/2204.02311} {{PaLM}: Scaling language modeling with pathways}.
\newblock \emph{arxiv:2204.02311}.

\bibitem[{Christiano et~al.(2017)Christiano, Leike, Brown, Martic, Legg, and Amodei}]{rlhf1}
Paul~F Christiano, Jan Leike, Tom Brown, Miljan Martic, Shane Legg, and Dario Amodei. 2017.
\newblock \href {https://proceedings.neurips.cc/paper_files/paper/2017/file/d5e2c0adad503c91f91df240d0cd4e49-Paper.pdf} {Deep reinforcement learning from human preferences}.
\newblock In \emph{Advances in Neural Information Processing Systems}, volume~30. Curran Associates, Inc.

\bibitem[{Craswell(2016)}]{rprec}
Nick Craswell. 2016.
\newblock \href {https://doi.org/10.1007/978-1-4899-7993-3_486-2} {\emph{R-Precision}}, pages 1--1. Springer New York, New York, NY.

\bibitem[{Guu et~al.(2020)Guu, Lee, Tung, Pasupat, and Chang}]{realm}
Kelvin Guu, Kenton Lee, Zora Tung, Panupong Pasupat, and Mingwei Chang. 2020.
\newblock \href {https://proceedings.mlr.press/v119/guu20a.html} {Retrieval augmented language model pre-training}.
\newblock In \emph{Proceedings of the 37th International Conference on Machine Learning}, volume 119 of \emph{Proceedings of Machine Learning Research}, pages 3929--3938. PMLR.

\bibitem[{Heinzerling and Inui(2021)}]{LMasKB2}
Benjamin Heinzerling and Kentaro Inui. 2021.
\newblock \href {https://doi.org/10.18653/v1/2021.eacl-main.153} {Language models as knowledge bases: On entity representations, storage capacity, and paraphrased queries}.
\newblock In \emph{Proceedings of the 16th Conference of the European Chapter of the Association for Computational Linguistics: Main Volume}, pages 1772--1791, Online. Association for Computational Linguistics.

\bibitem[{Jayaram~Subramanya et~al.(2019)Jayaram~Subramanya, Devvrit, Simhadri, Krishnawamy, and Kadekodi}]{diskann}
Suhas Jayaram~Subramanya, Fnu Devvrit, Harsha~Vardhan Simhadri, Ravishankar Krishnawamy, and Rohan Kadekodi. 2019.
\newblock \href {https://proceedings.neurips.cc/paper_files/paper/2019/file/09853c7fb1d3f8ee67a61b6bf4a7f8e6-Paper.pdf} {Diskann: Fast accurate billion-point nearest neighbor search on a single node}.
\newblock In \emph{Advances in Neural Information Processing Systems}, volume~32. Curran Associates, Inc.

\bibitem[{Johnson et~al.(2019)Johnson, Douze, and J{\'e}gou}]{faiss}
Jeff Johnson, Matthijs Douze, and Herv{\'e} J{\'e}gou. 2019.
\newblock \href {https://doi.org/10.1109/TBDATA.2019.2921572} {Billion-scale similarity search with {GPUs}}.
\newblock \emph{IEEE Transactions on Big Data}, 7(3):535--547.

\bibitem[{Karpukhin et~al.(2020)Karpukhin, Oguz, Min, Lewis, Wu, Edunov, Chen, and Yih}]{dpr}
Vladimir Karpukhin, Barlas Oguz, Sewon Min, Patrick Lewis, Ledell Wu, Sergey Edunov, Danqi Chen, and Wen-tau Yih. 2020.
\newblock \href {https://doi.org/10.18653/v1/2020.emnlp-main.550} {Dense passage retrieval for open-domain question answering}.
\newblock In \emph{Proceedings of the 2020 Conference on Empirical Methods in Natural Language Processing (EMNLP)}, pages 6769--6781, Online. Association for Computational Linguistics.

\bibitem[{Kwiatkowski et~al.(2019)Kwiatkowski, Palomaki, Redfield, Collins, Parikh, Alberti, Epstein, Polosukhin, Devlin, Lee, Toutanova, Jones, Kelcey, Chang, Dai, Uszkoreit, Le, and Petrov}]{nq}
Tom Kwiatkowski, Jennimaria Palomaki, Olivia Redfield, Michael Collins, Ankur Parikh, Chris Alberti, Danielle Epstein, Illia Polosukhin, Jacob Devlin, Kenton Lee, Kristina Toutanova, Llion Jones, Matthew Kelcey, Ming-Wei Chang, Andrew~M. Dai, Jakob Uszkoreit, Quoc Le, and Slav Petrov. 2019.
\newblock \href {https://doi.org/10.1162/tacl_a_00276} {Natural questions: A benchmark for question answering research}.
\newblock \emph{Transactions of the Association for Computational Linguistics}, 7:452--466.

\bibitem[{Lee(2023)}]{wizardvicuna}
June Lee. 2023.
\newblock {WizardVicunaLM}.
\newblock \url{https://github.com/melodysdreamj/WizardVicunaLM}.

\bibitem[{Lewis et~al.(2020{\natexlab{a}})Lewis, Liu, Goyal, Ghazvininejad, Mohamed, Levy, Stoyanov, and Zettlemoyer}]{bart}
Mike Lewis, Yinhan Liu, Naman Goyal, Marjan Ghazvininejad, Abdelrahman Mohamed, Omer Levy, Veselin Stoyanov, and Luke Zettlemoyer. 2020{\natexlab{a}}.
\newblock \href {https://doi.org/10.18653/v1/2020.acl-main.703} {{BART}: Denoising sequence-to-sequence pre-training for natural language generation, translation, and comprehension}.
\newblock In \emph{Proceedings of the 58th Annual Meeting of the Association for Computational Linguistics}, pages 7871--7880, Online. Association for Computational Linguistics.

\bibitem[{Lewis et~al.(2020{\natexlab{b}})Lewis, Perez, Piktus, Petroni, Karpukhin, Goyal, K\"{u}ttler, Lewis, Yih, Rockt\"{a}schel, Riedel, and Kiela}]{rag}
Patrick Lewis, Ethan Perez, Aleksandra Piktus, Fabio Petroni, Vladimir Karpukhin, Naman Goyal, Heinrich K\"{u}ttler, Mike Lewis, Wen-tau Yih, Tim Rockt\"{a}schel, Sebastian Riedel, and Douwe Kiela. 2020{\natexlab{b}}.
\newblock \href {https://proceedings.neurips.cc/paper_files/paper/2020/file/6b493230205f780e1bc26945df7481e5-Paper.pdf} {Retrieval-augmented generation for knowledge-intensive nlp tasks}.
\newblock In \emph{Advances in Neural Information Processing Systems}, volume~33, pages 9459--9474. Curran Associates, Inc.

\bibitem[{{LF Projects}(2023)}]{mlflow}
{LF Projects}. 2023.
\newblock {MLflow} -- a platform for the machine learning lifecycle.
\newblock \url{https://mlflow.org/}.

\bibitem[{Lin et~al.(2021)Lin, Ma, Lin, Yang, Pradeep, and Nogueira}]{pyserini}
Jimmy Lin, Xueguang Ma, Sheng-Chieh Lin, Jheng-Hong Yang, Ronak Pradeep, and Rodrigo Nogueira. 2021.
\newblock \href {https://doi.org/10.1145/3404835.3463238} {{Pyserini}: A {Python} toolkit for reproducible information retrieval research with sparse and dense representations}.
\newblock In \emph{Proceedings of the 44th Annual International ACM SIGIR Conference on Research and Development in Information Retrieval (SIGIR 2021)}, pages 2356--2362.

\bibitem[{Liska et~al.(2022)Liska, Kocisky, Gribovskaya, Terzi, Sezener, Agrawal, De~Masson~D'Autume, Scholtes, Zaheer, Young, Gilsenan-Mcmahon, Austin, Blunsom, and Lazaridou}]{streamingqa}
Adam Liska, Tomas Kocisky, Elena Gribovskaya, Tayfun Terzi, Eren Sezener, Devang Agrawal, Cyprien De~Masson~D'Autume, Tim Scholtes, Manzil Zaheer, Susannah Young, Ellen Gilsenan-Mcmahon, Sophia Austin, Phil Blunsom, and Angeliki Lazaridou. 2022.
\newblock \href {https://proceedings.mlr.press/v162/liska22a.html} {{S}treaming{QA}: A benchmark for adaptation to new knowledge over time in question answering models}.
\newblock In \emph{Proceedings of the 39th International Conference on Machine Learning}, volume 162 of \emph{Proceedings of Machine Learning Research}, pages 13604--13622. PMLR.

\bibitem[{Ma et~al.(2023)Ma, Gong, He, Zhao, and Duan}]{rewrite}
Xinbei Ma, Yeyun Gong, Pengcheng He, Hai Zhao, and Nan Duan. 2023.
\newblock \href {http://arxiv.org/abs/2305.14283} {Query rewriting for retrieval-augmented large language models}.
\newblock \emph{arXiv preprint arXiv:2305.14283}.

\bibitem[{Malkov and Yashunin(2020)}]{hnsw}
Yu~A. Malkov and D.~A. Yashunin. 2020.
\newblock \href {https://doi.org/10.1109/TPAMI.2018.2889473} {Efficient and robust approximate nearest neighbor search using hierarchical navigable small world graphs}.
\newblock \emph{IEEE Trans. Pattern Anal. Mach. Intell.}, 42(4):824--836.

\bibitem[{Mialon et~al.(2023)Mialon, Dess{\`\i}, Lomeli, Nalmpantis, Pasunuru, Raileanu, Rozi{\`e}re, Schick, Dwivedi-Yu, Celikyilmaz, Grave, LeCun, and Scialom}]{survey}
Gr{\'e}goire Mialon, Roberto Dess{\`\i}, Maria Lomeli, Christoforos Nalmpantis, Ram Pasunuru, Roberta Raileanu, Baptiste Rozi{\`e}re, Timo Schick, Jane Dwivedi-Yu, Asli Celikyilmaz, Edouard Grave, Yann LeCun, and Thomas Scialom. 2023.
\newblock \href {http://arxiv.org/abs/2302.07842} {Augmented language models: a survey}.
\newblock \emph{arXiv preprint arXiv:2302.07842}.

\bibitem[{Muennighoff et~al.(2023)Muennighoff, Tazi, Magne, and Reimers}]{mteb}
Niklas Muennighoff, Nouamane Tazi, Loic Magne, and Nils Reimers. 2023.
\newblock \href {https://aclanthology.org/2023.eacl-main.148} {{MTEB}: Massive text embedding benchmark}.
\newblock In \emph{Proceedings of the 17th Conference of the European Chapter of the Association for Computational Linguistics}, pages 2014--2037, Dubrovnik, Croatia. Association for Computational Linguistics.

\bibitem[{Nakano et~al.(2021)Nakano, Hilton, Balaji, Wu, Ouyang, Kim, Hesse, Jain, Kosaraju, Saunders et~al.}]{webgpt}
Reiichiro Nakano, Jacob Hilton, Suchir Balaji, Jeff Wu, Long Ouyang, Christina Kim, Christopher Hesse, Shantanu Jain, Vineet Kosaraju, William Saunders, et~al. 2021.
\newblock \href {http://arxiv.org/abs/2112.09332} {Webgpt: Browser-assisted question-answering with human feedback}.
\newblock \emph{arXiv preprint arXiv:2112.09332}.

\bibitem[{Ng et~al.(2023)Ng, Miyashita, Hoshi, Morioka, Torii, Kodama, and Deguchi}]{simplyretrieve}
Youyang Ng, Daisuke Miyashita, Yasuto Hoshi, Yasuhiro Morioka, Osamu Torii, Tomoya Kodama, and Jun Deguchi. 2023.
\newblock \href {http://arxiv.org/abs/2308.03983} {{SimplyRetrieve}: A private and lightweight retrieval-centric generative ai tool}.
\newblock \emph{arXiv preprint arXiv:2308.03983}.

\bibitem[{OpenAI(2023)}]{gpt4}
OpenAI. 2023.
\newblock \href {http://arxiv.org/abs/2303.08774} {{GPT-4} technical report}.
\newblock \emph{arXiv preprint arXiv:2303.08774}, abs/2303.08774.

\bibitem[{Petroni et~al.(2021)Petroni, Piktus, Fan, Lewis, Yazdani, De~Cao, Thorne, Jernite, Karpukhin, Maillard, Plachouras, Rockt{\"a}schel, and Riedel}]{kilt}
Fabio Petroni, Aleksandra Piktus, Angela Fan, Patrick Lewis, Majid Yazdani, Nicola De~Cao, James Thorne, Yacine Jernite, Vladimir Karpukhin, Jean Maillard, Vassilis Plachouras, Tim Rockt{\"a}schel, and Sebastian Riedel. 2021.
\newblock \href {https://doi.org/10.18653/v1/2021.naacl-main.200} {{KILT}: a benchmark for knowledge intensive language tasks}.
\newblock In \emph{Proceedings of the 2021 Conference of the North American Chapter of the Association for Computational Linguistics: Human Language Technologies}, pages 2523--2544, Online. Association for Computational Linguistics.

\bibitem[{Ram et~al.(2022)Ram, Bezalel, Zicher, Belinkov, Berant, and Globerson}]{ram2022you}
Ori Ram, Liat Bezalel, Adi Zicher, Yonatan Belinkov, Jonathan Berant, and Amir Globerson. 2022.
\newblock \href {http://arxiv.org/abs/2212.10380} {What are you token about? dense retrieval as distributions over the vocabulary}.
\newblock \emph{arXiv preprint arXiv:2212.10380}.

\bibitem[{Robertson and Zaragoza(2009)}]{bm25}
Stephen Robertson and Hugo Zaragoza. 2009.
\newblock \href {https://doi.org/10.1561/1500000019} {The probabilistic relevance framework: {BM25} and beyond}.
\newblock \emph{Found. Trends Inf. Retr.}, 3(4):333--389.

\bibitem[{Shi et~al.(2023)Shi, Min, Yasunaga, Seo, James, Lewis, Zettlemoyer, and Yih}]{replug}
Weijia Shi, Sewon Min, Michihiro Yasunaga, Minjoon Seo, Rich James, Mike Lewis, Luke Zettlemoyer, and Wen-tau Yih. 2023.
\newblock \href {http://arxiv.org/abs/2301.12652} {{RePlug}: Retrieval-augmented black-box language models}.
\newblock \emph{arXiv preprint arXiv:2301.12652}.

\bibitem[{Stiennon et~al.(2020)Stiennon, Ouyang, Wu, Ziegler, Lowe, Voss, Radford, Amodei, and Christiano}]{rlhf2}
Nisan Stiennon, Long Ouyang, Jeffrey Wu, Daniel Ziegler, Ryan Lowe, Chelsea Voss, Alec Radford, Dario Amodei, and Paul~F Christiano. 2020.
\newblock \href {https://proceedings.neurips.cc/paper_files/paper/2020/file/1f89885d556929e98d3ef9b86448f951-Paper.pdf} {Learning to summarize with human feedback}.
\newblock In \emph{Advances in Neural Information Processing Systems}, volume~33, pages 3008--3021. Curran Associates, Inc.

\bibitem[{Touvron et~al.(2023{\natexlab{a}})Touvron, Lavril, Izacard, Martinet, Lachaux, Lacroix, Rozi{\`e}re, Goyal, Hambro, Azhar, Rodriguez, Joulin, Grave, and Lample}]{llama}
Hugo Touvron, Thibaut Lavril, Gautier Izacard, Xavier Martinet, Marie-Anne Lachaux, Timoth{\'e}e Lacroix, Baptiste Rozi{\`e}re, Naman Goyal, Eric Hambro, Faisal Azhar, Aurelien Rodriguez, Armand Joulin, Edouard Grave, and Guillaume Lample. 2023{\natexlab{a}}.
\newblock \href {http://arxiv.org/abs/2302.13971} {{LLaMA}: Open and efficient foundation language models}.
\newblock \emph{arXiv preprint arXiv:2302.13971}.

\bibitem[{Touvron et~al.(2023{\natexlab{b}})Touvron, Martin, Stone, Albert, Almahairi, Babaei, Bashlykov, Batra, Bhargava, Bhosale et~al.}]{llama2}
Hugo Touvron, Louis Martin, Kevin Stone, Peter Albert, Amjad Almahairi, Yasmine Babaei, Nikolay Bashlykov, Soumya Batra, Prajjwal Bhargava, Shruti Bhosale, et~al. 2023{\natexlab{b}}.
\newblock \href {http://arxiv.org/abs/2307.09288} {Llama 2: Open foundation and fine-tuned chat models}.
\newblock \emph{arXiv preprint arXiv:2307.09288}.

\bibitem[{Wang et~al.(2022)Wang, Yang, Huang, Jiao, Yang, Jiang, Majumder, and Wei}]{e5}
Liang Wang, Nan Yang, Xiaolong Huang, Binxing Jiao, Linjun Yang, Daxin Jiang, Rangan Majumder, and Furu Wei. 2022.
\newblock \href {http://arxiv.org/abs/2212.03533} {Text embeddings by weakly-supervised contrastive pre-training}.
\newblock \emph{arXiv preprint arXiv:2212.03533}.

\bibitem[{Wolf et~al.(2020)Wolf, Debut, Sanh, Chaumond, Delangue, Moi, Cistac, Rault, Louf, Funtowicz, Davison, Shleifer, von Platen, Ma, Jernite, Plu, Xu, Le~Scao, Gugger, Drame, Lhoest, and Rush}]{huggingface}
Thomas Wolf, Lysandre Debut, Victor Sanh, Julien Chaumond, Clement Delangue, Anthony Moi, Pierric Cistac, Tim Rault, Remi Louf, Morgan Funtowicz, Joe Davison, Sam Shleifer, Patrick von Platen, Clara Ma, Yacine Jernite, Julien Plu, Canwen Xu, Teven Le~Scao, Sylvain Gugger, Mariama Drame, Quentin Lhoest, and Alexander Rush. 2020.
\newblock \href {https://doi.org/10.18653/v1/2020.emnlp-demos.6} {Transformers: State-of-the-art natural language processing}.
\newblock In \emph{Proceedings of the 2020 Conference on Empirical Methods in Natural Language Processing: System Demonstrations}, pages 38--45, Online. Association for Computational Linguistics.

\bibitem[{Xu et~al.(2023)Xu, Sun, Zheng, Geng, Zhao, Feng, Tao, and Jiang}]{wizardlm}
Can Xu, Qingfeng Sun, Kai Zheng, Xiubo Geng, Pu~Zhao, Jiazhan Feng, Chongyang Tao, and Daxin Jiang. 2023.
\newblock \href {http://arxiv.org/abs/2304.12244} {{WizardLM}: Empowering large language models to follow complex instructions}.
\newblock \emph{arXiv preprint arXiv:2304.12244}.

\bibitem[{Yao et~al.(2023)Yao, Zhao, Yu, Du, Shafran, Narasimhan, and Cao}]{react}
Shunyu Yao, Jeffrey Zhao, Dian Yu, Nan Du, Izhak Shafran, Karthik~R Narasimhan, and Yuan Cao. 2023.
\newblock \href {https://openreview.net/forum?id=WE_vluYUL-X} {{ReAct}: Synergizing reasoning and acting in language models}.
\newblock In \emph{The Eleventh International Conference on Learning Representations}.

\bibitem[{Zhou et~al.(2023)Zhou, Muresanu, Han, Paster, Pitis, Chan, and Ba}]{ape}
Yongchao Zhou, Andrei~Ioan Muresanu, Ziwen Han, Keiran Paster, Silviu Pitis, Harris Chan, and Jimmy Ba. 2023.
\newblock \href {https://openreview.net/forum?id=92gvk82DE-} {Large language models are human-level prompt engineers}.
\newblock In \emph{The Eleventh International Conference on Learning Representations}.

\end{thebibliography}
\bibliographystyle{acl_natbib}

\appendix
\section{Appendix}

\subsection{Computational Resources}

The evaluation experiments are conducted on an Ubuntu 20.04.6 server equipped with Intel(R) Xeon(R) Gold 6326 CPU at 2.90 GHz CPU cores, and one node with 4$\times$NVIDIA A100 Tensor Core GPU with 40 GB memory, and a RAID-5 array with a Dell(R) PERC H745 Front controller and KIOXIA(R) PM6-R SAS SSDs for storage.
The CUDA version is 12.2, the Python version is 3.9.16, the PyTorch version is 2.0.1, and the Transformers version is 4.29.2.

\begin{figure*}[h]
	\centering
	\includegraphics[width=1\textwidth]{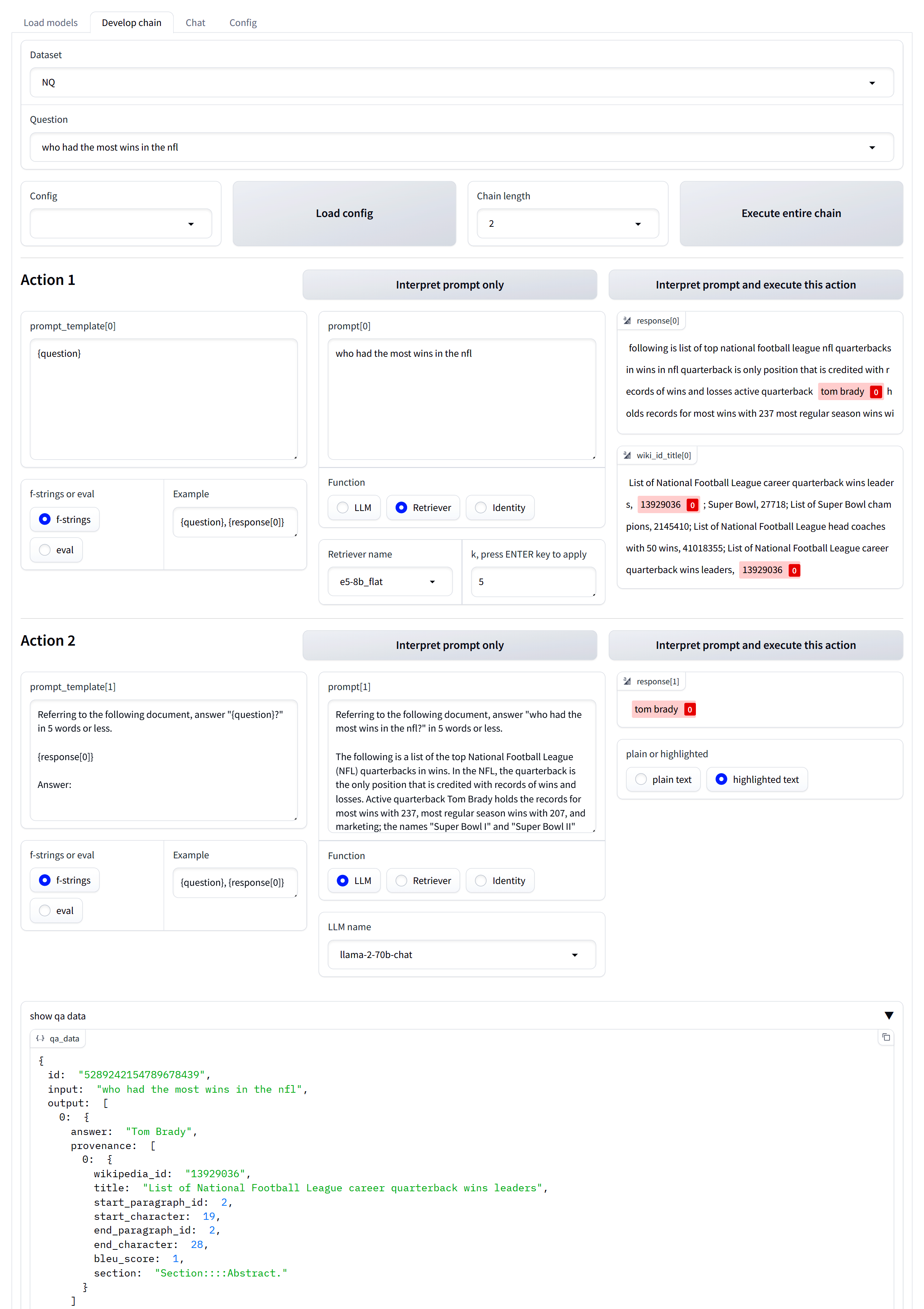}
	\caption{
		A screenshot of the \textit{Development chain} tab of \ralle.
		Developers can create tailored action chains comprising multiple actions of inference.
		For each action, developers can specify a prompt template, confirm the results of applying the template, and execute the action using the newly defined prompt, individually.
		Moreover, \rallez can highlight the gold answers within the retrieved documents or the output of the LLM, as well as highlight the Wikipedia IDs of successfully retrieved provenance.
	}
	\label{fig:dev_screen}
\end{figure*}

\clearpage

\subsection{Development Screen of \ralle}\label{sec:ralle_screen}

Figure~\ref{fig:dev_screen} shows the chain development screen\footnote{Please also review the \href{https://youtu.be/JYbm75qnfTg}{demonstration screencast}.}.
Developers can create an inference chain for an R-LLM on this \textit{Develop chain} tab.
One can choose a dataset and specify the desired chain length, which represents the total number of actions.
By default, there are two actions: retrieving with a retriever and generating with an LLM.

Prompt templates for each action can be defined using f-strings or eval functions in Python.
The results of applying the template can be confirmed without executing retrieval and generation.
The execution result can be viewed by clicking the \textit{Interpret prompt and execute this action} button.

The available action operators are \textit{LLM}, \textit{Retriever}, and \textit{Identity}.
\textit{LLM} generates text based on the given prompt.
\textit{Retriever} retrieves the top $k$ most relevant documents related to the input query.
And \textit{Identity} simply outputs the original prompt without employing a retriever or an LLM.

To execute the entire chain, click the \textit{Execute entire chain} button.
At the bottom of this tab, the selected question and its corresponding answer can be reviewed.
Also, \rallez enables to highlight the gold answers within the retrieved documents or the output of the LLM, as well as highlight the Wikipedia ID of successfully retrieved provenance.

\subsection{Additional Metric: has\_answer}\label{sec:has_answer}

\rallez also includes has\_answer percentage \citep[e.g.,][]{dpr} for short answers, a proxy metric to measure the proportion of questions that contain gold answers within the final output generated by an R-LLM.
By tracking this metric, developers can identify situations where the model generates responses that include gold answers but may be overlooked due to evaluation biases such as exact matching.
This information can help refine prompts to improve overall performance.

\subsection{Attempts to Build 3-action Chain}\label{sec:3_action}

According to Section~\ref{sec:kilt_downstream}, retrieval augmentation has a significant impact on performance in fact checking, open-domain QA for short answers, and slot filling tasks when comparing the closed-book and open-book settings of Llama2-70B.
In entity linking task (AY2, WnWi, and WnCw), however, our approach described in Section~\ref{sec:prompts} (retrieve, then output the top-1 retrieved Wikipedia title) may not be effective.

To improve the performance, we construct a \texttt{3-action} chain for AY2 dataset: (1) retrieve top-5 relevant documents, (2) explain the entity mention being questioned, and (3) predict the Wikipedia title based on the explanation and top-5 retrieved titles.
Additionally, we explore developing 3-action chains for T-REx and NQ datasets, which involves (1) retrieval, (2) question rewriting, and (3) answer generation.
Table~\ref{tab:prompt_template_3actions} shows the prompts used in 3-action chains.

Table~\ref{tab:3_action} shows the downstream performances with the \texttt{3-action} chains on AY2, NQ, and T-REx datasets.
While the 3-action chain outperforms the 2-action (retrieve-then-generate) chain on NQ dataset, it underperforms the 2-action accuracies on AY2 and T-REx datasets.
This suggests that the 3-action chains constructed specifically for these two datasets require further optimization.
However, the has\_answer value for AY2 (70.0\%) is higher than that of the 2-action chain (47.8\%), indicating that incorporating post-processing steps into the 3-action chain (thus to be 4-action chain) could potentially boost accuracy, particularly for AY2.

One of the benefits of our tool is that it allows for easy definition of such additional inference actions.
This means that developers can customize the chain to perform specific tasks beyond the default setting, giving them greater flexibility and control over their development.

\begin{table*}[t]
	\centering
	\renewcommand{\arraystretch}{1.2}
	\scalebox{0.65}
	{
		\begin{tabular}{l|ccccccccccc}
			\toprule
			\multicolumn{1}{c}{}                      & Fact Check.                        & \multicolumn{3}{c}{Entity Linking} & \multicolumn{2}{c}{Slot Filling}  & \multicolumn{4}{c}{Open Domain QA} & \multicolumn{1}{c}{Dial.}                                                                                                                                                                                                            \\
			\cmidrule(rl){2-2} \cmidrule(rl){3-5} \cmidrule(rl){6-7} \cmidrule(rl){8-11} \cmidrule(rl){12-12}
			\textit{Dataset}                          & \textbf{FEV}                       & \textbf{AY2}                       & \textbf{WnWi}                     & \textbf{WnCw}                      & \textbf{T-REx}                    & \textbf{zsRE}                      & \textbf{NQ}                       & \textbf{HoPo}                              & \textbf{TQA}                               & \textbf{ELI5} & \textbf{WoW}  \\
			\midrule
			\textit{Model / Metric}                   & \multicolumn{6}{c|}{Accuracy}      & \multicolumn{3}{c|}{Exact Match}   & \multicolumn{1}{c|}{RL}           & \multicolumn{1}{c}{F1}                                                                                                                                                                                                                                                    \\
			\midrule
			Llama2-70B (\small{\textit{closed-book}}) & 33.6 \scriptsize{(74.9)}           & 39.8 \scriptsize{(54.5)}           & 42.8 \scriptsize{(\textbf{53.8})} & 39.2 \scriptsize{(\textbf{55.7})}  & 28.5 \scriptsize{(40.5)}          & 11.3 \scriptsize{(13.6)}           & 19.6 \scriptsize{(37.4)}          & 13.9 \scriptsize{(25.1)}                   & 67.4 \scriptsize{(80.8)}                   & \textbf{23.0} & \textbf{13.3} \\
			\midrule
			RAG$^\diamondsuit$                        & \textbf{87.7}                      & \textbf{77.4}                      & \textbf{49.0}                     & \textbf{46.7}                      & \textbf{61.5}                     & \textbf{47.4}                      & \textbf{48.8}                     & 27.7                                       & 61.7                                       & 16.1          & \textbf{13.3} \\
			e5 + Llama2-70B                           & 49.9 \scriptsize{(\textbf{88.6})}  & 51.2 \scriptsize{(57.9)}           & 48.6 \scriptsize{(51.4)}          & 45.6 \scriptsize{(51.4)}           & 28.9 \scriptsize{(\textbf{49.2})} & 35.0  \scriptsize{(\textbf{43.2})} & 36.4 \scriptsize{(48.8)}          & \textbf{28.1} \scriptsize{(\textbf{35.8})} & \textbf{71.1} \scriptsize{(\textbf{83.9})} & 21.5          & 13.2          \\
			\quad \texttt{3-action}                   & -                                  & 24.4 \scriptsize{(\textbf{70.0})}  & -                                 & -                                  & 16.3 \scriptsize{(46.8)}          & -                                  & 36.9 \scriptsize{(\textbf{49.3})} & -                                          & -                                          & -             & -             \\
			\midrule
			\textit{Model / Metric}                   & \multicolumn{6}{c|}{KILT-Accuracy} & \multicolumn{3}{c|}{KILT-EM}       & \multicolumn{1}{c|}{KILT-RL}      & \multicolumn{1}{c}{KILT-F1}                                                                                                                                                                                                                                               \\
			\midrule
			RAG$^\diamondsuit$                        & \textbf{55.5}                      & \textbf{77.4}                      & \textbf{49.0}                     & \textbf{46.7}                      & \textbf{25.4}                     & \textbf{42.6}                      & \textbf{36.3}                     & 3.1                                        & 36.1                                       & \textbf{2.7}  & 7.5           \\
			e5 + Llama2-70B                           & 40.2  \scriptsize{(\textbf{71.2})} & 51.2 \scriptsize{(\textbf{51.2})}  & 48.6 \scriptsize{(\textbf{48.6})} & 45.5 \scriptsize{(\textbf{45.5})}  & 19.2 \scriptsize{(\textbf{29.7})} & 32.8 \scriptsize{(\textbf{40.4})}  & 27.7 \scriptsize{(36.3)}          & \textbf{11.3} \scriptsize{(\textbf{14.5})} & \textbf{42.8} \scriptsize{(\textbf{49.7})} & \textbf{2.7}  & \textbf{8.1}  \\
			\quad \texttt{3-action}                   & -                                  & 9.5 \scriptsize{(27.7)}            & -                                 & -                                  & 10.4 \scriptsize{(27.9)}          & -                                  & 28.0 \scriptsize{(\textbf{36.6})} & -                                          & -                                          & -             & -             \\
			\bottomrule
		\end{tabular}
	}
	\caption{
		Downstream performance of the \texttt{3-action} chain on KILT dev set along with baselines.
		The figures in parentheses represent has\_answer percentage, which corresponds to the proportion of questions with gold answers included in the final output of the LLM.
		$^\diamondsuit$: Results from \citet{kilt}.
	}
	\label{tab:3_action}
\end{table*}

\subsection{Details of Baseline Model in Open-Book Setting}\label{sec:rag_details}

As a baseline in open-book setting, we present the results of the Retrieval-Augmented Generation (RAG) model \citep{rag} shown in \citet{kilt}, which achieved strong performance in the KILT benchmark.
The RAG model comprises a bi-encoder retriever and a sequence-to-sequence generator (BART model \citep{bart}), both of which are trained end-to-end.
The total number of trainable parameters in the RAG model is approximately 626 million.
It is important to note that the RAG model was trained specifically for the KILT benchmark, whereas our chosen LLMs and constructed R-LLMs were not.

\begin{table*}[t]
	\centering
	\renewcommand{\arraystretch}{1.1}
	\scalebox{0.72}
	{
		\begin{tabular}{l|ccccccccccc}
			\toprule
			\multicolumn{1}{c}{}      & Fact Check.                   & \multicolumn{3}{c}{Entity Linking} & \multicolumn{2}{c}{Slot Filling} & \multicolumn{4}{c}{Open Domain QA} & \multicolumn{1}{c}{Dial.}                                                                                                                                                               \\
			\cmidrule(rl){2-2} \cmidrule(rl){3-5} \cmidrule(rl){6-7} \cmidrule(rl){8-11} \cmidrule(rl){12-12}
			\textit{Dataset}          & \textbf{FEV}                  & \textbf{AY2}                       & \textbf{WnWi}                    & \textbf{WnCw}                      & \textbf{T-REx}            & \textbf{zsRE}                     & \textbf{NQ}              & \textbf{HoPo}            & \textbf{TQA}                      & \textbf{ELI5} & \textbf{WoW}  \\
			\midrule
			\textit{Model / Metric}   & \multicolumn{6}{c|}{Accuracy} & \multicolumn{3}{c|}{Exact Match}   & \multicolumn{1}{c|}{RL}          & \multicolumn{1}{c}{F1}                                                                                                                                                                                                       \\
			\midrule
			BART-large$^\diamondsuit$ & \textbf{80.7}                 & \textbf{86.6}                      & \textbf{47.9}                    & \textbf{48.0}                      & \textbf{43.8}             & 3.0                               & \textbf{26.2}            & \textbf{16.9}            & 32.5                              & 22.7          & \textbf{13.8} \\
			W-Vicuna-13B              & 0.0 \scriptsize{(58.4)}       & 0.1 \scriptsize{(52.2)}            & 2.0 \scriptsize{(44.9)}          & 0.0 \scriptsize{(48.1)}            & 17.9 \scriptsize{(33.0)}  & 5.9 \scriptsize{(8.5)}            & 6.2 \scriptsize{(27.4)}  & 1.7 \scriptsize{(17.1)}  & 20.0 \scriptsize{(64.5)}          & 22.7          & 12.7          \\
			Llama2-13B                & 26.3 \scriptsize{(50.7)}      & 34.6 \scriptsize{(47.5)}           & 35.0 \scriptsize{(42.8)}         & 28.5 \scriptsize{(41.3)}           & 26.9  \scriptsize{(36.7)} & 7.8 \scriptsize{(9.9)}            & 11.5 \scriptsize{(29.1)} & 8.3 \scriptsize{(20.3)}  & 43.0 \scriptsize{(70.2)}          & \textbf{27.6} & 13.0          \\
			Llama2-70B                & 33.6 \scriptsize{(74.9)}      & 39.8 \scriptsize{(54.5)}           & 42.8 \scriptsize{(53.8)}         & 39.2 \scriptsize{(55.7)}           & 28.5 \scriptsize{(40.5)}  & \textbf{11.3} \scriptsize{(13.6)} & 19.6 \scriptsize{(37.4)} & 13.9 \scriptsize{(25.1)} & \textbf{67.4} \scriptsize{(80.8)} & 23.0          & 13.3          \\
			\bottomrule
		\end{tabular}
	}
	\caption{
		Downstream performance on KILT development set in a \textit{closed-book} setting (generation without retrieval).
		Following \citet{kilt}, we report the results of typical metrics for each dataset, with bold indicating the best result.
		The figures in parentheses represent has\_answer percentage, which corresponds to the proportion of questions with gold answers included in the final output of the LLM.
		$^\diamondsuit$: Results from \citet{kilt}.
	}
	\label{tab:kilt_downstream_closed}
\end{table*}

\subsection{KILT Downstream Performances in Closed-Book Setting}\label{sec:kilt_downstream_closed}

Table~\ref{tab:kilt_downstream_closed} summarizes the KILT downstream results in a closed-book setting.
The baseline (BART-large) model has been fine-tuned on the KILT datasets, while our chosen LLMs have not.
Despite this, the LLMs demonstrate superior performance compared to the baseline on several datasets.

Specifically, the Llama2-70B model outperforms the BART baseline on the zsRE and TQA datasets, and the Llama2-13B model outperforms the baseline on the ELI5 dataset.
This suggests that the parametric knowledge embedded in the LLMs and their capacity for text generation can be leveraged effectively for knowledge-intensive tasks, even zero-shot setting.
Nevertheless, as described in Section~\ref{sec:kilt_downstream}, retrieval augmentation can enhance the performance on downstream tasks, except the ELI5 dataset.
We also present the closed-book performances of several LLMs on the development set of NQ dataset in Table \ref{tab:closed_nq}.

\newcolumntype{Y}{&gt;{\centering\arraybackslash}X}
\begin{table*}[h]
	\centering
	\renewcommand{\arraystretch}{1.15}
	\scalebox{1.0}
	{
		\begin{tabular}{lcccc}
			\toprule
			\multicolumn{5}{c}{\textit{NQ}}                                          \\
			\midrule
			Model Name       & EM            & has\_answer   & f1            & sec/Q \\
			\midrule
			Llama-2-70b-chat & 19.6          & 37.4          & 36.8          & 2.254 \\
			Llama-2-13b-chat & 11.5          & 29.1          & 28.1          & 1.179 \\
			StableBeluga2    & 16.2          & \textbf{40.9} & 35.5          & 2.858 \\
			gpt-3.5-turbo    & \textbf{25.4} & 38.9          & \textbf{41.1} & -     \\
			\bottomrule
		\end{tabular}
	}
	\caption{
		Accuracies on NQ dev set in a closed-book setting.
		For gpt-3.5-turbo (version 0613), the accuracy was calculated excluding five questions out of 2,837 questions in the NQ development set that were deemed inappropriate prompts by OpenAI and were not processed.
	}
	\label{tab:closed_nq}
\end{table*}

\subsection{Additional Results for Retrieval Performance}\label{sec:recall}

Table~\ref{tab:recall} presents the recall@5 of the retrievers used in our experiments.
Note that even though m-e5 outperforms e5 on the MTEB Retrieval task (shown in Table \ref{tab:retrievers}), e5 still demonstrates superior performance compared to m-e5 in terms of both R-precision (shown in Table~\ref{tab:kilt_retrieval}) and recall@5.

\begin{table*}[t]
	\centering
	\renewcommand{\arraystretch}{1.1}
	\scalebox{0.65}
	{
		\begin{tabular}{l|ccccccccccc|l}
			\toprule
			\multicolumn{1}{c}{}                & Fact Check.   & \multicolumn{3}{c}{Entity Linking} & \multicolumn{2}{c}{Slot Filling} & \multicolumn{4}{c}{Open Domain QA} & \multicolumn{1}{c}{Dial.} & \multicolumn{1}{c}{}                                                                                                                \\
			\cmidrule(rl){2-2} \cmidrule(rl){3-5} \cmidrule(rl){6-7} \cmidrule(rl){8-11} \cmidrule(rl){12-12}
			\textit{Dataset}            & \textbf{FEV}  & \textbf{AY2}                       & \textbf{WnWi}                    & \textbf{WnCw}                      & \textbf{T-REx}            & \textbf{zsRE}        & \textbf{NQ}   & \textbf{HoPo} & \textbf{TQA}  & \textbf{ELI5} & \textbf{WoW}  & Avg.                         \\
			\midrule
			% Model                       & \multicolumn{9}{c}{R-Precision} &                                    & \multicolumn{1}{c}{}               & \multicolumn{1}{c}{}                                                                                                                                                                                       \\
			% \midrule
			\textit{Model}                & \multicolumn{11}{c}{\large{Recall@5}} & \multicolumn{1}{c}{}                                                                                                                                                                                                                                                         \\
			\midrule
			RAG$^\diamondsuit$                  & 76.1          & \textbf{77.5}                      & 49.0                             & 46.7                               & 33.7                      & 73.1                 & 65.5          & 12.3          & 56.9          & \textbf{27.3} & 66.6          & 53.1                         \\
			BM25                                & 74.2          & 28.8                               & 34.7                             & 30.6                               & 42.7                      & 74.7                 & 42.5          & 22.8          & 48.7          & 12.3          & 45.1          & 41.6                         \\
			\quad $-$ \texttt{REWRITE-EL}       & \sgr{74.2}    & 7.6 \scriptsize{($-$21.2)}         & 3.1 \scriptsize{($-$31.6)}       & 5.9* \scriptsize{($-$24.7)}        & \sgr{42.7}                & \sgr{74.7}           & \sgr{42.5}    & \sgr{22.8}    & \sgr{48.7}    & \sgr{12.3}    & \sgr{45.1}    & 34.5 \scriptsize{($-$7.1)}   \\
			m-e5 (Flat)                         & \textbf{91.0} & 58.5                               & 60.6                             & 62.2                               & \textbf{53.1}             & 87.0                 & 69.5          & 40.4          & \textbf{65.4} & 19.1          & 75.0          & 62.0                         \\
			\quad $-$ \texttt{REWRITE-EL}       & \sgr{91.0}    & 7.8 \scriptsize{($-$50.7)}         & 3.8 \scriptsize{($-$56.8)}       & 5.5 \scriptsize{($-$56.7)}         & \sgr{53.1}                & \sgr{87.0}           & \sgr{69.5}    & \sgr{40.4}    & \sgr{65.4}    & \sgr{19.1}    & \sgr{75.0}    & 47.1 \scriptsize{($-$14.9)}  \\
			m-e5 (HNSW) $-$ \texttt{REWRITE-EL} & 63.2          & 4.9                                & 3.5                              & 2.6                                & 26.0                      & 48.2                 & 55.6          & 14.1          & 48.7          & 14.6          & 66.8          & 31.7                         \\
			e5 (Flat)                           & 90.6          & 66.1                               & \textbf{63.3}                    & \textbf{66.7}                      & 52.1                      & \textbf{87.2}        & \textbf{71.6} & \textbf{40.9} & \textbf{65.4} & 21.3          & \textbf{75.3} & \textbf{63.7}                \\
			\quad $-$ \texttt{REWRITE-EL}       & \sgr{90.6}    & 7.6 \scriptsize{($-$58.5)}         & 3.4 \scriptsize{($-$59.9)}       & 4.8 \scriptsize{($-$61.9)}         & \sgr{52.1}                & \sgr{87.2}           & \sgr{71.6}    & \sgr{40.9}    & \sgr{65.4}    & \sgr{21.3}    & \sgr{75.3}    & 47.3 \scriptsize{($-$16.4)}  \\
			e5 (HNSW)                           & 74.7          & 49.6                               & 50.7                             & 50.8                               & 26.7                      & 55.9                 & 65.2          & 19.3          & 58.6          & 16.0          & 70.9          & 48.9                         \\
			\quad $-$ \texttt{REWRITE-EL}       & \sgr{74.7}    & 6.0 \scriptsize{($-$43.6)}         & 3.3 \scriptsize{($-$47.4)}       & 3.3 \scriptsize{($-$47.5)}         & \sgr{26.7}                & \sgr{55.9}           & \sgr{65.2}    & \sgr{19.3}    & \sgr{58.6}    & \sgr{16.0}    & \sgr{70.9}    & 36.4  \scriptsize{($-$12.5)} \\
			e5 (DiskANN)                        & 86.6          & 57.1                               & 58.3                             & 60.9                               & 42.1                      & 78.7                 & 70.7          & 34.7          & 64.6          & 20.8          & 75.0          & 59.0                         \\
			\quad $-$ \texttt{REWRITE-EL}       & \sgr{86.6}    & 7.4 \scriptsize{($-$49.7)}         & 3.3 \scriptsize{($-$55.0)}       & 3.6 \scriptsize{($-$57.3)}         & \sgr{42.1}                & \sgr{78.7}           & \sgr{70.7}    & \sgr{34.7}    & \sgr{64.6}    & \sgr{20.8}    & \sgr{75.0}    & 44.3 \scriptsize{($-$14.7)}  \\
			\bottomrule
		\end{tabular}
	}
	\caption{
		Retrieval performances (recall@5) on KILT dev set.
		\textit{Avg.} refers to macro-average of the scores in each dataset.
		Bold indicates the best result.
		The figures shown in gray are copied from the column above because they do not change based on the given setting.
		$^\diamondsuit$: Results from \citet{kilt}.
		*: BM25 (without \texttt{REWRITE-EL}) failed with long queries (45 out of 5,599 questions) in WnCw.
	}
	\label{tab:recall}
\end{table*}

\newcommand{\avedev}[2]{{#1}\scriptsize{$\pm${#2}}}

\begin{table*}[t]
	\centering
	\renewcommand{\arraystretch}{1.15}
	\scalebox{0.61}
	{
		\begin{tabular}{l|ccccccccccc|c}
			\toprule
			\multicolumn{1}{c}{}                & Fact Check.                                                                      & \multicolumn{3}{c}{Entity Linking} & \multicolumn{2}{c}{Slot Filling} & \multicolumn{4}{c}{Open Domain QA} & \multicolumn{1}{c}{Dial.} & \multicolumn{1}{c}{}                                                                                                                   \\
			\cmidrule(rl){2-2} \cmidrule(rl){3-5} \cmidrule(rl){6-7} \cmidrule(rl){8-11} \cmidrule(rl){12-12}
			\textit{Tasks}                      & \textbf{FEV}                                                                     & \textbf{AY2}                       & \textbf{WnWi}                    & \textbf{WnCw}                      & \textbf{T-REx}            & \textbf{zsRE}        & \textbf{NQ}       & \textbf{HoPo}      & \textbf{TQA}       & \textbf{ELI5}       & \textbf{WoW}        & Avg.  \\
			\midrule
			\textit{Models}                     & \multicolumn{10}{c}{Completion in Closed-Book Setting (in seconds per question)} & \multicolumn{1}{c}{}               & \multicolumn{1}{c}{}                                                                                                                                                                                                                       \\
			\midrule
			W-Vicuna-13B                        & 1.565                                                                            & 13.040                             & 10.870                           & 9.793                              & 0.983                     & 1.142                & 2.165             & 1.969              & 1.414              & 22.820              & 7.122               & 6.626 \\
			Llama2-13B                          & 0.625                                                                            & 1.077                              & 1.036                            & 1.201                              & 0.940                     & 0.913                & 1.270             & 1.185              & 1.014              & 40.100              & 9.522               & 5.353 \\
			Llama2-70B                          & 1.765                                                                            & 2.936                              & 2.745                            & 2.618                              & 1.953                     & 2.031                & 2.285             & 2.188              & 1.877              & 42.500              & 11.100              & 6.727 \\
			\midrule
			\multicolumn{13}{c}{Retrieval + Generation (in seconds per question)}                                                                                                                                                                                                                                                                                                                                    \\
			\midrule
			e5 + W-Vicuna-13B                   & 1.529                                                                            & 1.310                              & 1.368                            & 1.158                              & 1.192                     & 1.453                & 2.595             & 1.945              & 1.734              & 15.480              & 10.850              & 3.692 \\
			e5 + Llama2-13B                     & 1.084                                                                            & 1.165                              & 1.209                            & 1.046                              & 1.300                     & 1.407                & 1.284             & 1.975              & 9.830              & 32.48               & 16.76               & 6.322 \\
			BM25 + Llama2-70B                   & 1.841                                                                            & 0.008                              & 0.009                            & 0.008                              & 2.015                     & 2.296                & 2.206             & 2.344              & 2.249              & 15.020              & 12.010              & 3.637 \\
			e5 + Llama2-70B                     & 1.926                                                                            & 0.133                              & 0.131                            & 0.135                              & 2.135                     & 2.424                & 2.419             & 2.346              & 2.238              & 16.030              & 11.810              & 3.793 \\
			e5 (top-2) + Llama2-70B             & 1.544                                                                            & 0.133                              & 0.131                            & 0.135                              & 1.661                     & 1.908                & 1.994             & 1.833              & 1.759              & 15.120              & 10.820              & 3.367 \\
			e5 (top-10) + Llama2-70B            & 2.811                                                                            & 0.133                              & 0.131                            & 0.135                              & 2.951                     & 3.276                & -                 & 13.900             & 14.400             & 35.070              & 24.100              & -     \\
			e5 (DiskANN) + Llama2-70B           & 1.803                                                                            & 0.044                              & 0.044                            & 0.043                              & 2.009                     & 2.281                & 2.166             & 2.247              & 2.116              & 15.780              & 11.370              & 3.628 \\
			e5 + Llama2-70B (\texttt{3-action}) & -                                                                                & 25.41                              & -                                & -                                  & 4.993                     & -                    & 16.320            & -                  & -                  & -                   & -                           \\
			\midrule
			\multicolumn{13}{c}{Retrieval (in seconds per question)}                                                                                                                                                                                                                                                                                                                                                 \\
			\midrule
			BM25                                & 0.038                                                                            & 0.008                              & 0.009                            & 0.008                              & 0.018                     & 0.013                & 0.052             & 0.105              & 0.086              & 0.136               & 0.857               & 0.121 \\
			BM25 (without \texttt{REWRITE-EL})  & 0.038                                                                            & 5.700                              & 4.531                            & 5.440                              & 0.018                     & 0.013                & 0.052             & 0.105              & 0.086              & 0.136               & 0.857               & 1.543 \\
			m-e5 (Flat)                         & 0.174                                                                            & 0.164                              & 0.166                            & 0.176                              & 0.187                     & 0.165                & 0.194             & 0.156              & 0.176              & 0.177               & 0.165               & 0.173 \\
			m-e5 (HNSW)                         & 0.008                                                                            & 0.013                              & 0.013                            & 0.015                              & 0.008                     & 0.009                & 0.009             & 0.009              & 0.009              & 0.011               & 0.010               & 0.010 \\
			e5 (Flat)                           & 0.177                                                                            & 0.168                              & 0.172                            & 0.159                              & 0.201                     & 0.170                & 0.171             & 0.146              & 0.155              & 0.174               & 0.165               & 0.169 \\
			e5 (HNSW)                           & 0.008                                                                            & 0.008                              & 0.008                            & 0.008                              & 0.008                     & 0.008                & 0.008             & 0.008              & 0.008              & 0.008               & 0.009               & 0.008 \\
			e5 (DiskANN)                        & 0.018                                                                            & 0.020                              & 0.038                            & 0.020                              & 0.020                     & 0.030                & 0.020             & 0.021              & 0.019              & 0.019               & 0.021               & 0.022 \\
			\midrule
			\multicolumn{13}{c}{Mean Query Length (tokens)}                                                                                                                                                                                                                                                                                                                                                          \\
			\midrule
			                                    & \avedev{11.1}{4.0}                                                               & \avedev{357.9}{149.0}              & \avedev{331.5}{113.5}            & \avedev{505.2}{31.1}               & \avedev{7.5}{2.5}         & \avedev{7.6}{2.3}    & \avedev{9.9}{2.1} & \avedev{19.5}{6.6} & \avedev{17.9}{8.9} & \avedev{21.0}{10.7} & \avedev{86.3}{58.0} &       \\
			\bottomrule
		\end{tabular}
	}
	\caption{
		Execution time (in seconds per question) in \ralle.
		\textit{Avg.} refers to macro-average of the times in each task.
		The mean query length and its standard deviation (shown as $\pm$ after the value) are also displayed, which were calculated using the e5 tokenizer.
	}
	\label{tab:details_speed}
\end{table*}

\subsection{Details of Speed Analysis}\label{sec:details_speed}

Table~\ref{tab:details_speed} presents the details of speed analysis on KILT development set.
The search speed of BM25 (without \texttt{REWRITE-EL}) decreases as the total number of words in a query increases.
In contrast, for dense vector search, the search speed remains relatively constant regardless of the size of the query due to the fixed dimensionality of the embedding vectors.

According to Table~\ref{tab:details_speed}, the execution times required for generation with an LLM is longer than the times required for retrieval, particularly when generating lengthy responses such as ELI5 and WoW.
Therefore, it may seem counterintuitive that the advantages of ANNS used in vector search are not fully realized in terms of execution time of R-LLMs.
However, as previously discussed in Section \ref{sec:speed}, DiskANN requires less memory compared to other vector search algorithms, which means that using such algorithm can actually help conserve computational resources for R-LLM.
% This tradeoff between memory usage and downstream accuracy can be carefully considered when selecting an ANNS algorithm for R-LLM using \ralle.

We observe that Llama2-13B requires more time to process each question compared to Llama2-70B.
Upon further analysis, we discovered that the Llama2-13B model occasionally produced nonsensical responses such as multiple newline characters (``\textbackslash n''), partially due to the limitations of our prompts.
% These issues highlight the importance of thorough experimentation when using these models.

\subsection{Model Information}\label{sec:model_info}

As shown in Table \ref{tab:model_info}, we utilize several open-source models from Hugging Face, specifically their officially released versions.
We load the distributed models in 8-bit precision by default except Llama2-70B model (in 4-bit) using Hugging Face Accelerate\footnote{\url{https://huggingface.co/docs/accelerate/index}} library.

\newcommand{\smallurl}[1]{\small{\url{#1}}} % tcolorbox

\newcolumntype{Y}{&gt;{\centering\arraybackslash}X}
\begin{table*}[h]
	\centering
	\renewcommand{\arraystretch}{1.15}
	\scalebox{0.67}
	{
		\begin{tabular}{lrccl}
			\toprule
			\multicolumn{5}{c}{\large{\textit{Language Model}}}                                                                                                       \\
			\midrule
			Model Name                             & \multicolumn{1}{c}{Size} & max len. & emb dim & URL                                                              \\
			\midrule
			wizard-vicuna-13b \citep{wizardvicuna} & 13,015,864,320           & 2,048    & -       & \smallurl{https://huggingface.co/junelee/wizard-vicuna-13b}      \\
			Llama-2-13b-chat \citep{llama2}        & 13,015,864,320           & 4,096    & -       & \smallurl{https://huggingface.co/meta-llama/Llama-2-13b-chat}    \\
			Llama-2-70b-chat \citep{llama2}        & 68,976,653,312           & 4,096    & -       & \smallurl{https://huggingface.co/meta-llama/Llama-2-70b-chat}    \\
			StableBeluga2                          & 70B                      & 4,096    & -       & \smallurl{https://huggingface.co/stabilityai/StableBeluga2}      \\
			\midrule
			\multicolumn{5}{c}{\large{\textit{Retriever}}}                                                                                                            \\
			\midrule
			multilingual-e5-large                  & 559,890,946              & 514      & 1,024   & \smallurl{https://huggingface.co/intfloat/multilingual-e5-large} \\
			e5-large-v2 \citep{e5}                 & 335,142,400              & 512      & 1,024   & \smallurl{https://huggingface.co/intfloat/e5-large-v2}           \\
			\bottomrule
		\end{tabular}
	}
	\caption{
		Hugging Face links of the models used in our evaluation.
		\textit{Size} refers to the total number of effective parameters of each model.
		\textit{max len.} refers to the maximum token length of model input.
	}
	\label{tab:model_info}
\end{table*}

\subsection{Using Custom Datasets}\label{sec:custom_datasets}

In addition to utilizing KILT datasets, \rallez enables developers to develop and evaluate R-LLMs on their own QA datasets and corpora.
To use the custom datasets with \ralle, you will need to perform the following preprocessing:

\begin{itemize}
    \item Prepare your corpus as a TSV file containing the document IDs, texts, and titles.
    \item Create a JSONL file for your QA dataset. The format should look like this: \{\texttt{"id": "", "input": "", "output": }[\{\texttt{"answer": "", "provenance": }[\{\texttt{"wikipedia\_id": "", "title": ""}\}]\}]\}, where ``input'' represents a question.
\end{itemize}

See our repo for more detailed instructions: \url{https://github.com/yhoshi3/RaLLe}.

\subsection{The Prompts used in the Evaluation}\label{sec:our_prompts}

Table~\ref{tab:prompt_template} summarizes the prompts used in our experiment.
\textit{Open-book} indicates retrieve-then-generate setting.
The queries used for retrieval are the raw questions without any rewriting, except for the \texttt{REWRITE-EL} settings of AY2, WnWi, and WnCw.

\textit{Closed-book} indicates that an LLM answers to the given question without retrieval.
Although these prompts have been our established best practices, we recognize that there may be opportunities for improvement (see also Section \ref{sec:limitations}).

\newcommand{\nlarrow}{\scalebox{0.5}{$\textcolor{gray}{\hookleftarrow}$}} % New Line Arrow ←
\newcommand{\promptbox}[1]{\raisebox{-\height}{\begin{tcolorbox}[opacityframe=0, enhanced jigsaw, enlarge top by=-4mm, top=1mm, left=1mm] {#1} \end{tcolorbox}}} % tcolorbox
\newcommand{\sq}{\textquotesingle} % straight single quotation mark '
\newcommand{\fstringon}{\hspace{2mm}\raisebox{-6.3pt}{\includegraphics[height=17pt]{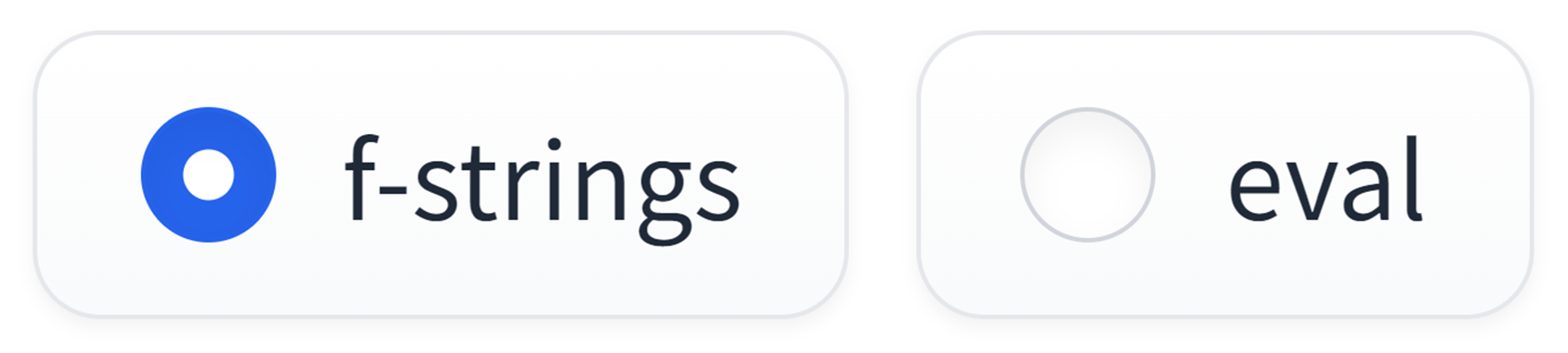}}} % insert f-string checked button
\newcommand{\evalon}{\hspace{2mm}\raisebox{-6.3pt}{\includegraphics[height=17pt]{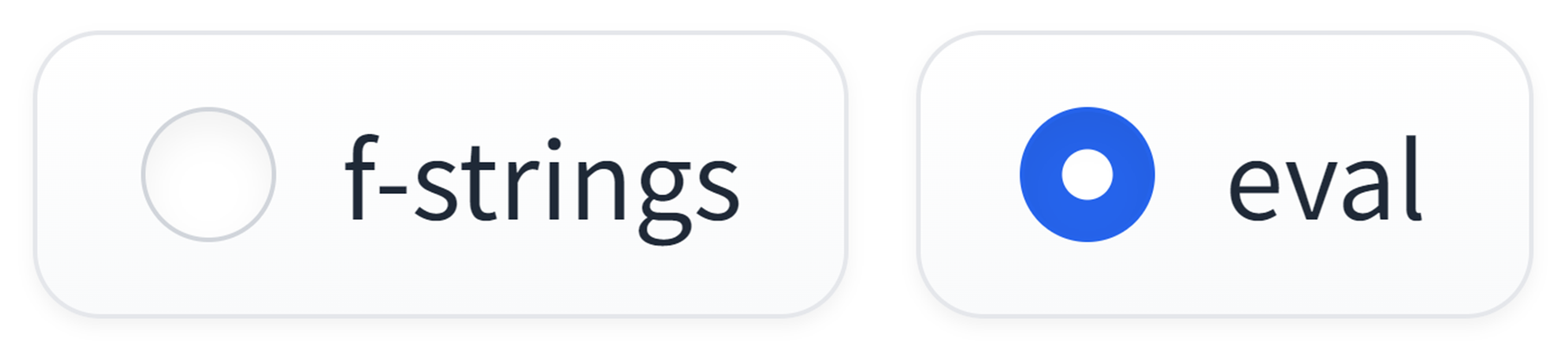}}}

\newcommand{\ttsplit}[1]{\texttt{\seqsplit{#1}}} % texttt and split long text
\newcommand{\bs}{\textbackslash} % texttt and split long text

\newcommand{\promptboxtiny}[1]{\raisebox{-\height}{\begin{tcolorbox}[opacityframe=0, enhanced jigsaw, enlarge top by=-3mm, enlarge bottom by=3mm, enlarge left by=-1mm, enlarge right by=5mm, top=1mm, bottom=1mm, left=1mm, right=0mm, fontupper=\tiny] {#1} \end{tcolorbox}}} % box for each action

\newcommand{\chainbox}[1]{\raisebox{-\height}{\begin{tcolorbox}[opacityframe=0, enhanced jigsaw, enlarge top by=-4mm, enlarge bottom by=0mm, enlarge left by=-1mm, enlarge right by=-1mm, top=1mm, bottom=-3mm, left=0mm, right=-4mm, colback=white] {#1} \end{tcolorbox}}} % box for each chain

\newcolumntype{Y}{&gt;{\centering\arraybackslash}X} %centering
\begin{table*}[h]
	\centering
	\renewcommand{\arraystretch}{1.15}
	% \scalebox{0.8}
	{
		\begin{tabular}{p{7.5cm}|p{7.5cm}}
			\toprule
			\multicolumn{1}{c|}{\textit{Open-book}} & \multicolumn{1}{c}{\textit{Closed-book}}                                                                                                                                                                                                                                                                  \\
			\midrule
			% FEVER
			\multicolumn{2}{c}{\qquad \textit{FEVER} \fstringon}                                                                                                                                                                                                                                                                                                \\
			\midrule
			\chainbox{
			Action 1: Retriever                                                                                                                                                                                                                                                                                                                                 \\
			\promptboxtiny{\ttsplit{\{question\}}}                                                                                                                                                                                                                                                                                                              \\
			Action 2: LLM                                                                                                                                                                                                                                                                                                                                       \\
				\promptboxtiny{
			\texttt{\{response[0]\}} \nlarrow                                                                                                                                                                                                                                                                                                                   \\
			\texttt{} \nlarrow                                                                                                                                                                                                                                                                                                                                  \\
			\texttt{Answer IN ONE WORD if the document SUPPORTS or REFUTES "\{question\}".} \nlarrow                                                                                                                                                                                                                                                            \\
			\texttt{} \nlarrow                                                                                                                                                                                                                                                                                                                                  \\
					\texttt{Answer: }
				}
			}
			                                        &
			\chainbox{
			Action 1: LLM                                                                                                                                                                                                                                                                                                                                       \\
				\promptboxtiny{
			\texttt{Answer IN ONE WORD if your knowledge SUPPORTS or REFUTES "\{question\}".} \nlarrow                                                                                                                                                                                                                                                          \\
			\texttt{} \nlarrow                                                                                                                                                                                                                                                                                                                                  \\
					\texttt{Answer: }
				}
			}                                                                                                                                                                                                                                                                                                                                                   \\
			\midrule
			% AY2
			\multicolumn{2}{c}{\qquad \textit{AY2} \evalon}                                                                                                                                                                                                                                                                                                     \\
			\midrule
			\chainbox{
			Action 1: Retriever                                                                                                                                                                                                                                                                                                                                 \\
				\promptboxtiny{
			\texttt{\sq What is "\sq  + \sq \{\}\sq .format(question).split(                                                                                                                                                                                                                                                                                    \\
						\sq [START\_ENT]\sq )[1].split(\sq [END\_ENT]\sq )[0][1:-1] + \sq " ?\sq }
			}                                                                                                                                                                                                                                                                                                                                                   \\
			Action 2: Identity                                                                                                                                                                                                                                                                                                                                  \\
				\promptboxtiny{
					\texttt{\sq \{\}\sq .format(wiki\_id\_title[0]).split(\sq ; \sq )[0].split(\sq , \sq )[0]}
				}
			}
			                                        &
			\chainbox{
			Action 1: LLM                                                                                                                                                                                                                                                                                                                                       \\
				\promptboxtiny{
			\texttt{\sq What is the most relevant Wikipedia title to the entity "\sq  + \sq \{\}\sq .format(question).split(\sq [START\_ENT] \sq )[1].split(\sq  [END\_ENT]\sq )[0] + \sq " in the context of "\sq  + \sq \{\}\sq .format(question).split(\sq [START\_ENT]\sq )[0][-100:] + \sq \{\}\sq .format(question).split(\sq [START\_ENT]\sq )[1].split( \\
			\sq [END\_ENT]\sq )[0] + \sq \{\}\sq .format(question).split(                                                                                                                                                                                                                                                                                       \\
						\sq [END\_ENT]\sq )[1][:100] + \sq \sq \sq ..."?\bs n\bs nPlease answer only the Wikipedia title.\bs n\bs nAnswer: \sq \sq \sq }
				}
			}                                                                                                                                                                                                                                                                                                                                                   \\
			\midrule
			% WnWi
			\multicolumn{2}{c}{\qquad \textit{WnWi} \evalon}                                                                                                                                                                                                                                                                                                    \\
			\midrule
			\chainbox{
			Action 1: Retriever                                                                                                                                                                                                                                                                                                                                 \\
				\promptboxtiny{
			\texttt{\sq What is "\sq  + \sq \{\}\sq .format(question).split(                                                                                                                                                                                                                                                                                    \\
						\sq [START\_ENT]\sq )[1].split(\sq [END\_ENT]\sq )[0][1:-1] + \sq " ?\sq }
			}                                                                                                                                                                                                                                                                                                                                                   \\
			Action 2: Identity                                                                                                                                                                                                                                                                                                                                  \\
				\promptboxtiny{
					\texttt{\sq \{\}\sq .format(wiki\_id\_title[0]).split(\sq ; \sq )[0].split(\sq , \sq )[0]}
				}
			}
			                                        &
			\chainbox{
			Action 1: LLM                                                                                                                                                                                                                                                                                                                                       \\
				\promptboxtiny{
			\texttt{\sq What is the most relevant Wikipedia title to the entity "\sq  + \sq \{\}\sq .format(question).split(\sq [START\_ENT] \sq )[1].split(\sq  [END\_ENT]\sq )[0] + \sq " in the context of "\sq  + \sq \{\}\sq .format(question).split(\sq [START\_ENT]\sq )[0][-100:] + \sq \{\}\sq .format(question).split(\sq [START\_ENT]\sq )[1].split( \\
			\sq [END\_ENT]\sq )[0] + \sq \{\}\sq .format(question).split(                                                                                                                                                                                                                                                                                       \\
						\sq [END\_ENT]\sq )[1][:100] + \sq \sq \sq ..."?\bs n\bs nPlease answer only the Wikipedia title.\bs n\bs nAnswer: \sq \sq \sq }
				}
			}                                                                                                                                                                                                                                                                                                                                                   \\
			\midrule
			% WnCw
			\multicolumn{2}{c}{\qquad \textit{WnCw} \evalon}                                                                                                                                                                                                                                                                                                    \\
			\midrule
			\chainbox{
			Action 1: Retriever                                                                                                                                                                                                                                                                                                                                 \\
				\promptboxtiny{
			\texttt{\sq What is "\sq  + \sq \{\}\sq .format(question).split(                                                                                                                                                                                                                                                                                    \\
						\sq [START\_ENT]\sq )[1].split(\sq [END\_ENT]\sq )[0][1:-1] + \sq " ?\sq }
			}                                                                                                                                                                                                                                                                                                                                                   \\
			Action 2: Identity                                                                                                                                                                                                                                                                                                                                  \\
				\promptboxtiny{
					\texttt{\sq \{\}\sq .format(wiki\_id\_title[0]).split(\sq ; \sq )[0].split(\sq , \sq )[0]}
				}
			}
			                                        &
			\chainbox{
			Action 1: LLM                                                                                                                                                                                                                                                                                                                                       \\
				\promptboxtiny{
			\texttt{\sq What is the most relevant Wikipedia title to the entity "\sq  + \sq \{\}\sq .format(question).split(\sq [START\_ENT] \sq )[1].split(\sq  [END\_ENT]\sq )[0] + \sq " in the context of "\sq  + \sq \{\}\sq .format(question).split(\sq [START\_ENT]\sq )[0][-100:] + \sq \{\}\sq .format(question).split(\sq [START\_ENT]\sq )[1].split( \\
			\sq [END\_ENT]\sq )[0] + \sq \{\}\sq .format(question).split(                                                                                                                                                                                                                                                                                       \\
						\sq [END\_ENT]\sq )[1][:100] + \sq \sq \sq ..."?\bs n\bs nPlease answer only the Wikipedia title.\bs n\bs nAnswer: \sq \sq \sq }
				}
			}                                                                                                                                                                                                                                                                                                                                                   \\
			\midrule
			\multicolumn{2}{r}{\textit{Continued on next page...}}                                                                                                                                                                                                                                                                                              \\
		\end{tabular}
	}
	\caption{
		Prompt templates used in our experiments.
		The hook-left arrows $\textcolor{gray}{\hookleftarrow}$ refers to new line.
		Note that \rallez supports f-strings and eval() function in Python.
	}
	\label{tab:prompt_template}
\end{table*}

\newcolumntype{Y}{&gt;{\centering\arraybackslash}X}
\begin{table*}[h]
	\centering
	\renewcommand{\arraystretch}{1.15}
	% \scalebox{0.8}
	{
		\begin{tabular}{p{7.5cm}|p{7.5cm}}
			\multicolumn{2}{c}{Table~\ref{tab:prompt_template} \textit{-- continued from previous page.}}     \\
			\midrule
			\multicolumn{1}{c|}{\textit{Open-book}} & \multicolumn{1}{c}{\textit{Closed-book}}                \\
			\midrule
			% T-REx
			\multicolumn{2}{c}{\textit{T-REx}}                                                                \\
			\midrule
			\chainbox{
			Action 1: Retriever (\texttt{f-strings})                                                          \\
				\promptboxtiny{
					\texttt{\{question\}}
				}
			\\
			Action 2: LLM (\texttt{eval()})                                                                   \\
				\promptboxtiny{
					\texttt{\sq \sq \sq Referring to the following document, answer "what is the \sq \sq \sq  + \sq \{\}\sq .format(question).split(\sq [SEP]\sq )[1] + \sq  of \sq  + \sq \{\}\sq .format(question).split(\sq [SEP]\sq )[0] + \sq \sq \sq ?" in 5 words or less.\bs n\bs n\sq \sq \sq  + \sq \{\}\sq .format(response[0]) + \sq \sq \sq \bs n\bs n\sq \sq \sq  + \sq \{\}\sq .format(question).split(\sq [SEP]\sq )[1] + \sq : \sq }
				}
			}
			                                        &
			\chainbox{
			Action 1: LLM (\texttt{eval()})                                                                   \\
				\promptboxtiny{
					\texttt{\sq What is the \sq  + \sq "\sq  + \sq \{\}\sq .format(question).split(\sq [SEP] \sq )[1] + \sq " of "\sq  + \sq \{\}\sq .format(question).split(\sq  [SEP]\sq )[0] + \sq "\sq  + \sq \sq \sq  in 5 words or less?\bs n\bs n\sq \sq \sq  + \sq \{\}\sq .format(question).split(\sq [SEP] \sq )[1] + \sq : \sq }
				}
			}                                                                                                 \\
			\midrule
			% zsRE
			\multicolumn{2}{c}{\textit{zsRE}}                                                                 \\
			\midrule
			\chainbox{
			Action 1: Retriever (\texttt{f-strings})                                                          \\
				\promptboxtiny{
					\texttt{\{question\}}
				}
			\\
			Action 2: LLM (\texttt{eval()})                                                                   \\
				\promptboxtiny{
			\texttt{Referring to the following document, answer "\{question\}?" in 5 words or less.} \nlarrow \\
			\texttt{} \nlarrow                                                                                \\
			\texttt{\{response[0]\}} \nlarrow                                                                 \\
			\texttt{} \nlarrow                                                                                \\
					\texttt{Answer: }
				}
			}
			                                        &
			\chainbox{
			Action 1: LLM (\texttt{eval()})                                                                   \\
				\promptboxtiny{
					\texttt{\sq Tell me the \sq  + \sq "\sq  + \sq \{\}\sq .format(question).split(\sq [SEP] \sq )[1] + \sq " of "\sq  + \sq \{\}\sq .format(question).split(\sq  [SEP]\sq )[0] + \sq "\sq  + \sq \sq \sq  in 5 words or less.\bs n\bs n\sq \sq \sq  + \sq \{\}\sq .format(question).split(\sq [SEP] \sq )[1] + \sq : \sq }
				}
			}                                                                                                 \\

			\midrule
			% NQ
			\multicolumn{2}{c}{\qquad \textit{NQ} \fstringon}                                                 \\
			\midrule
			\chainbox{
			Action 1: Retriever                                                                               \\
				\promptboxtiny{
					\texttt{\{question\}}
				}
			\\
			Action 2: LLM                                                                                     \\
				\promptboxtiny{
			\texttt{Referring to the following document, answer "\{question\}?" in 5 words or less.} \nlarrow \\
			\texttt{} \nlarrow                                                                                \\
			\texttt{\{response[0]\}} \nlarrow                                                                 \\
			\texttt{} \nlarrow                                                                                \\
					\texttt{Answer: }
				}
			}
			                                        &
			\chainbox{
			Action 1: LLM                                                                                     \\
				\promptboxtiny{
			\texttt{Answer \sq \{question\}?\sq  in 5 words or less.} \nlarrow                                \\
			\texttt{} \nlarrow                                                                                \\
					\texttt{Answer: }
				}
			}                                                                                                 \\
			\midrule
			% HoPo
			\multicolumn{2}{c}{\qquad \textit{HoPo} \fstringon}                                               \\
			\midrule
			\chainbox{
			Action 1: Retriever                                                                               \\
				\promptboxtiny{
					\texttt{\{question\}}
				}
			\\
			Action 2: LLM                                                                                     \\
				\promptboxtiny{
			\texttt{Referring to the following document, answer "\{question\}?" in 5 words or less.} \nlarrow \\
			\texttt{} \nlarrow                                                                                \\
			\texttt{\{response[0]\}} \nlarrow                                                                 \\
			\texttt{} \nlarrow                                                                                \\
					\texttt{Answer: }
				}
			}
			                                        &
			\chainbox{
			Action 1: LLM                                                                                     \\
				\promptboxtiny{
			\texttt{Answer \sq \{question\}?\sq  in 5 words or less.} \nlarrow                                \\
			\texttt{} \nlarrow                                                                                \\
					\texttt{Answer: }
				}
			}                                                                                                 \\
			\midrule
			\multicolumn{2}{r}{\textit{Continued on next page...}}                                            \\
		\end{tabular}
	}
	% \label{tab:prompt_template}
\end{table*}

\newcolumntype{Y}{&gt;{\centering\arraybackslash}X}
\begin{table*}[h]
	\centering
	\renewcommand{\arraystretch}{1.15}
	% \scalebox{0.8}
	{
		\begin{tabular}{p{7.5cm}|p{7.5cm}}
			\multicolumn{2}{c}{Table~\ref{tab:prompt_template} \textit{-- continued from previous page.}}                          \\
			\midrule
			\multicolumn{1}{c|}{\textit{Open-book}} & \multicolumn{1}{c}{\textit{Closed-book}}                                     \\
			\midrule
			% TQA
			\multicolumn{2}{c}{\qquad \textit{TQA} \fstringon}                                                                     \\
			\midrule
			\chainbox{
			Action 1: Retriever                                                                                                    \\
				\promptboxtiny{
					\texttt{\{question\}}
				}
			\\
			Action 2: LLM                                                                                                          \\
				\promptboxtiny{
			\texttt{Referring to the following document, answer "\{question\}?" in 5 words or less.} \nlarrow                      \\
			\texttt{} \nlarrow                                                                                                     \\
			\texttt{\{response[0]\}} \nlarrow                                                                                      \\
			\texttt{} \nlarrow                                                                                                     \\
					\texttt{Answer: }
				}
			}
			                                        &
			\chainbox{
			Action 1: LLM                                                                                                          \\
				\promptboxtiny{
			\texttt{Answer \sq \{question\}\sq  in 5 words or less.} \nlarrow                                                      \\
			\texttt{} \nlarrow                                                                                                     \\
					\texttt{Answer: }
				}
			}                                                                                                                      \\
			\midrule
			% ELI5
			\multicolumn{2}{c}{\qquad \textit{ELI5} \fstringon}                                                                    \\
			\midrule
			\chainbox{
			Action 1: Retriever                                                                                                    \\
				\promptboxtiny{
					\texttt{\{question\}}
				}
			\\
			Action 2: LLM                                                                                                          \\
				\promptboxtiny{
			\texttt{Referring to the following document, answer "\{question\}".} \nlarrow                                          \\
			\texttt{} \nlarrow                                                                                                     \\
			\texttt{\{response[0]\}} \nlarrow                                                                                      \\
			\texttt{} \nlarrow                                                                                                     \\
			\texttt{Explain the following questions as if I were five years old.} \nlarrow                                         \\
			\texttt{\{question\}} \nlarrow                                                                                         \\
			\texttt{} \nlarrow                                                                                                     \\
					\texttt{Answer: }
				}
			}
			                                        &
			\chainbox{
			Action 1: LLM                                                                                                          \\
				\promptboxtiny{
			\texttt{Explain \sq \{question\}\sq  as if I were five years old.} \nlarrow                                            \\
			\texttt{} \nlarrow                                                                                                     \\
					\texttt{Answer: }
				}
			}                                                                                                                      \\
			\midrule
			% WoW
			\multicolumn{2}{c}{\qquad \textit{WoW} \fstringon}                                                                     \\
			\midrule
			\chainbox{
			Action 1: Retriever                                                                                                    \\
				\promptboxtiny{
					\texttt{\{question\}}
				}
			\\
			Action 2: LLM                                                                                                          \\
				\promptboxtiny{
			\texttt{Referring to the following document, output a short and informative reply to the conversation.} \nlarrow       \\
			\texttt{} \nlarrow                                                                                                     \\
			\texttt{\{response[0]\}} \nlarrow                                                                                      \\
			\texttt{} \nlarrow                                                                                                     \\
			\texttt{Referring to the above document, output a short and informative reply to the following conversation.} \nlarrow \\
			\texttt{} \nlarrow                                                                                                     \\
			\texttt{This conversation ends on your turn.} \nlarrow                                                                 \\
			\texttt{} \nlarrow                                                                                                     \\
			\texttt{\{question\}} \nlarrow                                                                                         \\
			\texttt{} \nlarrow                                                                                                     \\
			\texttt{Informative and short answer:} \nlarrow                                                                        \\
			\texttt{} \nlarrow                                                                                                     \\
					\texttt{}
				}
			}
			                                        &
			\chainbox{
			Action 1: LLM                                                                                                          \\
				\promptboxtiny{
			\texttt{Output a short and informative reply to the conversation. This conversation ends on your turn.} \nlarrow       \\
			\texttt{} \nlarrow                                                                                                     \\
			\texttt{\{question\}} \nlarrow                                                                                         \\
			\texttt{} \nlarrow                                                                                                     \\
					\texttt{Informative and short answer: }
				}
			}                                                                                                                      \\
			\bottomrule
		\end{tabular}
	}
	% \label{tab:prompt_template}
\end{table*}

\newcolumntype{Y}{&gt;{\centering\arraybackslash}X}
\begin{table*}[h]
	\centering
	\renewcommand{\arraystretch}{1.15}
	% \scalebox{0.8}
	{
		\begin{tabular}{l}
			\toprule
			\multicolumn{1}{c}{\qquad \textit{AY2} \evalon}                                                     \\
			\midrule
			\chainbox{
			Action 1: Retriever                                                                                 \\
				\promptboxtiny{
					\texttt{\sq What is "\sq  + \sq \{\}\sq .format(question).split(\sq [START\_ENT] \sq )[1].split(\sq  [END\_ENT]\sq )[0] + \sq " in the context of "\sq  + \sq \{\}\sq .format(question).split(\sq [START\_ENT]\sq )[0][-100:] + \sq \{\}\sq .format(question).split(\sq [START\_ENT]\sq )[1].split(\sq [END\_ENT]\sq )[0] + \sq \{\}\sq .format(question).split(\sq [END\_ENT]\sq )[1][:100] + \sq ..."?\sq }
				}
			\\
			Action 2: LLM                                                                                       \\
				\promptboxtiny{
					\texttt{\sq What is "\sq  + \sq \{\}\sq .format(question).split(\sq [START\_ENT] \sq )[1].split(\sq  [END\_ENT]\sq )[0] + \sq " in the context of "\sq  + \sq \{\}\sq .format(question).split(\sq [START\_ENT]\sq )[0][-100:] + \sq \{\}\sq .format(question).split(\sq [START\_ENT]\sq )[1].split(\sq [END\_ENT]\sq )[0] + \sq \{\}\sq .format(question).split(\sq [END\_ENT]\sq )[1][:100] + \sq ..."?\bs nAnswer in a short and conc sentence.\sq  + \sq \sq \sq \bs n\bs nAnswer:\bs n\sq \sq \sq }
				}
			\\
			Action 3: LLM                                                                                       \\
				\promptboxtiny{
					\texttt{\sq Please select the most appropriate title for the word "\sq  + \sq \{\}\sq .format(question).split(\sq [START\_ENT] \sq )[1].split(\sq  [END\_ENT]\sq )[0] + \sq " based on the given Description.\sq  + \sq \sq \sq \bs nIf none of these titles suit your needs, please suggest a possible alternative title.\sq \sq \sq  + \sq \sq \sq \bs Titles: \bs n\sq \sq \sq  + \sq  / \sq .join([titleid.split(\sq ,\sq )[0] for titleid in \sq \{\}\sq .format(wiki\_id\_title[0]).split(\sq ; \sq )]) + \sq \sq \sq \bs n\bs nDescription:\bs n\sq \sq \sq  + \sq \{\}\sq .format(response[1]) + \sq \sq \sq \bs n\bs nWikipedia Title:\bs n\sq \sq \sq }
				}
			}                                                                                                   \\
			\midrule
			\multicolumn{1}{c}{\qquad \textit{T-REx} \fstringon}                                                \\
			\midrule
			\chainbox{
			Action 1: Retriever                                                                                 \\
				\promptboxtiny{
					\texttt{\{question\}}
				}
			\\
			Action 2: LLM                                                                                       \\
				\promptboxtiny{
			\texttt{Formulate a question that asks [SEP] in the following sentence:} \nlarrow                   \\
			\texttt{\sq \{question\}\sq } \nlarrow                                                              \\
			\texttt{} \nlarrow                                                                                  \\
					\texttt{Generated question: }
				}
			\\
			Action 3: LLM                                                                                       \\
				\promptboxtiny{
			\texttt{\{response[0]\}} \nlarrow                                                                   \\
			\texttt{} \nlarrow                                                                                  \\
			\texttt{Referring to the document above, answer "\{response[1]\}" in 5 words or less.} \nlarrow     \\
			\texttt{} \nlarrow                                                                                  \\
					\texttt{Answer: }
				}
			}                                                                                                   \\
			\midrule
			\multicolumn{1}{c}{\qquad \textit{NQ} \fstringon}                                                   \\
			\midrule
			\chainbox{
			Action 1: Retriever                                                                                 \\
				\promptboxtiny{
					\texttt{\{question\}}
				}
			\\
			Action 2: LLM                                                                                       \\
				\promptboxtiny{
			\texttt{Please rewrite the following question clearly.} \nlarrow                                    \\
			\texttt{} \nlarrow                                                                                  \\
			\texttt{\{question\}?} \nlarrow                                                                     \\
			\texttt{} \nlarrow                                                                                  \\
					\texttt{Rewritten question: }
				}
			\\
			Action 3: LLM                                                                                       \\
				\promptboxtiny{
			\texttt{Referring to the following document, answer "\{response[1]\}" in 5 words or less.} \nlarrow \\
			\texttt{} \nlarrow                                                                                  \\
			\texttt{\{response[0]\}} \nlarrow                                                                   \\
			\texttt{} \nlarrow                                                                                  \\
					\texttt{Answer: }
				}
			}                                                                                                   \\
			\bottomrule
		\end{tabular}
	}
	\caption{
		Prompt templates used in \texttt{3-action} chains.
	}
	\label{tab:prompt_template_3actions}
\end{table*}

\end{document}